\documentclass[runningheads]{llncs}
\usepackage{graphicx}
\usepackage{amsmath,amssymb} % define this before the line numbering.
\usepackage{color}
\usepackage[width=122mm,left=12mm,paperwidth=146mm,height=193mm,top=12mm,paperheight=217mm]{geometry}
\usepackage{booktabs}
\usepackage{multirow}
\usepackage{xr}
\usepackage[pagebackref=true,breaklinks=true,letterpaper=true,colorlinks,bookmarks=false]{hyperref}

\newcommand{\etal}{\mbox{\emph{et al.\ }}}
\newcommand{\ra}[1]{\renewcommand{\arraystretch}{#1}}

\begin{document}
\pagestyle{headings}
\mainmatter

\title{Stacked U-Nets:  A No-Frills Approach to Natural Image Segmentation}

\titlerunning{Stacked U-Nets:  A No-Frills Approach to Natural Image Segmentation}

\authorrunning{Sohil Shah, Pallabi Ghosh, Larry S. Davis and Tom Goldstein}

\author{Sohil Shah, Pallabi Ghosh, Larry S. Davis \and Tom Goldstein}
\institute{University of Maryland, College Park, MD USA.\\
\email{\{sohilas,lsdavis\}@umd.edu,\{pallabig,tomg\}@cs.umd.edu}}

\maketitle
\begin{abstract}
Many imaging tasks require global information about all pixels in an image. Conventional bottom-up classification networks globalize information by decreasing resolution; features are pooled and down-sampled into a single output. But for semantic segmentation and object detection tasks, a network must provide higher-resolution pixel-level outputs. To globalize information while preserving resolution, many researchers propose the inclusion of sophisticated auxiliary blocks, but these come at the cost of a considerable increase in network size and computational cost. This paper proposes stacked u-nets (SUNets), which iteratively combine features from different resolution scales while maintaining resolution. SUNets leverage the information globalization power of u-nets in a deeper network architectures that is capable of handling the complexity of natural images. SUNets perform extremely well on semantic segmentation tasks using a small number of parameters. The code is available at \url{https://github.com/shahsohil/sunets}.
\end{abstract} 

\section{Introduction}
Semantic segmentation methods decompose an image into groups of pixels, each representing a common object class.  While the output of a segmentation contains object labels assigned at the local (pixel) level, each label much have a global {\em field of view}; each such label depends on global information about the image, such as textures, colors, and object boundaries that may span large chunks of the image.  Simple image classification algorithms consolidate global information by successively pooling features until the final output is a single label containing information from the entire image.  In contrast, segmentation methods must output a full-resolution labeled image (rather than a single label).  Thus, a successful segmentation method must address this key question: how can we learn long-distance contextual information while at the same time retaining high spatial resolution at the output for identifying small objects and sharp boundaries?

For natural image processing, most research has answered this question using one of two approaches. One approach is to use very few pooling layers, thus maintaining resolution (although methods may still require a small number of deconvolution layers~\cite{lin2017refinenet,noh2015learning,fu2017stacked,jegou2017one,ghiasi2016laplacian}).  Large fields of view are achieved using dilated convolutions, which span large regions. By maintaining resolution at each layer, this approach preserves substantial amounts of signal about smaller and less salient objects. However, this is achieved at the cost of computationally expensive and memory exhaustive training/inference. The second related approach is to produce auxiliary context aggregation blocks~\cite{krahenbuhl2011efficient,chen2016deeplab,liang2015semantic,chandra2016fast,chandra2017dense,zheng2015conditional,schwing2015fully,yu2015multi,zhao2017pyramid,chen2017rethinking} that contain features at different distance scales, and then merge these blocks to produce a final segmentation. This category includes many well-known techniques such as dense CRF~\cite{krahenbuhl2011efficient} (conditional random fields) and spatial pyramid pooling~\cite{chen2016deeplab}. 

These approaches suffer from the following challenges:
\begin{enumerate}
\item  Deconvolutional (i.e., encoder-decoder) architectures perform significant nonlinear computation at low resolutions, but do very little processing of high-resolution features. 
During the convolution/encoding stage, pooling layers can move information over large distances, but information about small objects is often lost.  During the deconvolution/decoding stage, high- and low-resolution features are merged to produce upsampled feature maps.  However, the high-resolution inputs to deconvolutional layers come from relatively shallow layers that do not effectively encode semantic information.
\item Image classification networks are parameter heavy ($44.5$M parameters for ResNet-101), and segmentation methods built on top of these classification networks are often even more burdensome.  For example, on top of the resnet-101 architecture, PSPNet~\cite{zhao2017pyramid} uses $22$M {\em additional} parameters for context aggregation, while the ASPP and Cascade versions of the Deeplab network utilize $14.5$M~\cite{chen2016deeplab} and $40$M~\cite{chen2017rethinking} {\em additional} parameters, respectively. 
\end{enumerate}

A popular and simple approach to segmentation is u-nets, which perform a chain of convolutional/downsampling operations, followed by a chain of deconvolutional/upsampling layers that see information from both low- and high-resolution scales.  These u-net architectures are state-of-the art for medical image segmentation~\cite{ronneberger2015u}, but they do not perform well when confronted with the complex color profiles, lighting effects, perspectives, and occlusions present in natural images.

We expand the power of u-nets by stacking u-net blocks into deep architectures.  This addresses the two challenges discussed above:  As data passes through multiple u-net blocks, high-resolution features are mixed with low-resolution context information and processed through many layers to produce informative high-resolution features.  Furthermore, stacked U-net models require fewer feature maps per layer than conventional architectures, and thus achieve higher performance with far fewer parameters. Our smallest model exceeds the performance of ResNet-101 on the PASCAL VOC 2012 semantic segmentation task by $4.5\%$ mIoU, while having $\sim 7\times$ fewer parameters.

\section{Related Work} \label{related}
Many models~\cite{chen2017rethinking,zhao2017pyramid,peng2017large,chen2016deeplab,wang2017understanding,ghiasi2016laplacian,lin2017refinenet,fu2017stacked} have boosted the performance of semantic segmentation networks. These gains are mainly attributed to the use of pre-trained models, dilated convolutional layers~\cite{chen2016deeplab,yu2015multi} and fully convolutional architectures (DCNN)~\cite{long2015fully}. These works employ a range of strategies to tap contextual information, which fall into three major categories.

{\bf Context Aggregation Modules:} These architectures place a special module on top of a pre-trained network that integrates context information at different distance scales. 
The development of fast and efficient algorithm for DenseCRF~\cite{krahenbuhl2011efficient} led to the development of numerous algorithms~\cite{chen2016deeplab,liang2015semantic,chandra2016fast,chandra2017dense} incorporating it on top of the output belief map. Moreover, the joint training of CRF and CNN parameters was made possible by~\cite{zheng2015conditional,schwing2015fully}. In~\cite{yu2015multi}, context information was integrated by processing a belief map using a cascade of dilated layers operating at progressively increasing dilation rates, and~\cite{wang2017understanding} proposed a hybrid dilation convolution framework to alleviate gridding artifacts. ParseNet~\cite{liu2015parsenet} exploits image-level feature information at each layer to learn global contextual information. In contrast,~\cite{zhao2017pyramid,chen2017rethinking,chen2016deeplab} realized substantial performance improvements by employing parallel layers of spatial pyramid pooling. The work~\cite{zhao2017pyramid} spatially pools output feature maps at different scales, while~\cite{chen2017rethinking,chen2016deeplab} advocates applying dilated convolution at varying dilation rates.

{\bf Image Pyramid:} The networks proposed in \cite{lin2016efficient,chen2016attention} learn context information by simultaneously processing inputs at different scales and merging the output from all scales. An attention mechanism was used to perform fusion of output maps in~\cite{chen2016attention}, while~\cite{lin2016efficient} concatenates all the feature maps produced by blocks of parallel layers, each learned exclusively for differently scaled inputs. Recently,~\cite{dai2017deformable} developed a deformable network that adaptively determines an object's scale and accordingly adjusts the receptive field size of each activation function. On the other hand,~\cite{singh2017analysis} proposed a new training paradigm for object detection networks that trains each object instance only using the proposals closest to the ground truth scale.

{\bf Encoder-Decoder:} These models consist of an encoder network and one or many blocks of decoder layers. The decoder fine-tunes the pixel-level labels by merging the contextual information from feature maps learned at all the intermediate layers. Usually, a popular bottom-up pre-trained classification network such as ResNet~\cite{he2016deep}, VGG~\cite{simonyan2014very} or DenseNet~\cite{huang2017densely} serves as an encoder model. U-net~\cite{ronneberger2015u} popularly employs skip connections between an encoder and its corresponding decoding layers. On a similar note, the decoder in Segnet~\cite{badrinarayanan2017segnet} upsamples the lower resolution maps by reusing the pooling indices from the encoder.  Deconvolution layers were stacked in \cite{noh2015learning,fu2017stacked,jegou2017one}, whereas~\cite{ghiasi2016laplacian} uses a Laplacian pyramid reconstruction network to selectively refine the low resolution maps. Refinenet~\cite{lin2017refinenet} employs sophisticated decoder modules at each scale on top of the ResNet encoder, while \cite{deeplabv3xception} utilizes a simple two-level decoding of feature maps from the Xception network~\cite{chollet2017xception}. In short, the structure in~\cite{deeplabv3xception,ghiasi2016laplacian} is a hybrid of decoding and context aggregation modules. Any task that requires extracting multi-scale information from inputs can benefit from an encoder-decoder structure. The recent works on object detection~\cite{shrivastava2016beyond,lin2017feature} also utilize this structure.

\subsection{Use of Pre-Trained Nets}
Many of the networks described above make extensive use of image classification networks that were pre-trained for other purposes.  ResNet employs an identity mapping~\cite{he2016identity} which, along with batch-normalization layers, facilitates efficient learning of very deep models. VGG was popular before the advent of ResNet. Although parameter heavy, much fundamental work on segmentation (FCN~\cite{long2015fully}, dilated nets~\cite{yu2015multi}, u-nets~\cite{ronneberger2015u} and CRF~\cite{zheng2015conditional}) was built on VGG.   All these architectures share common origins in that they were designed for the ImageNet competition and features are processed bottom-up. This prototype works well when the network has to identify only a single object without exact pixel localization. However, when extended to localization tasks such as segmentation and object detection, it is not clear whether the complete potential of these networks has been properly tapped. Recent work on object detection~\cite{singh2017analysis} also echoes a similar concern.

\section{U-Nets Revisited}
The original u-net architecture was introduced in \cite{ronneberger2015u}, and produced almost perfect segmentation of cells in biomedical images using very little training data. The structure of u-nets makes it possible to capture context information at multiple scales and propagate them to the higher resolution layers. These higher order features have enabled u-nets to outperform previous deep models~\cite{long2015fully} on various tasks including semantic segmentation~\cite{milletari2016v,cciccek20163d,jegou2017one}, depth-fusion~\cite{Riegler2017OctNetFusion}, image translation~\cite{pix2pix2017} and human-pose estimation~\cite{newell2016stacked}. 
Moreover, driven by the initial success of u-nets, many recent works on semantic segmentation~\cite{lin2017refinenet,ghiasi2016laplacian,badrinarayanan2017segnet,noh2015learning,fu2017stacked,jegou2017one} and object detection~\cite{shrivastava2016beyond,lin2017feature} also propose an encoder-decoder deep architecture. 

The u-net architecture evenly distributes its capacity among the encoder and decoder modules. Moreover, the complete network can be trained in an end-to-end setting. In contrast, the more recent architectures reviewed in Section \ref{related} do not equally distribute the processing of top-down and bottom-up features. Since these architectures are built on top of pre-trained feature extractors~\cite{simonyan2014very,he2016deep,chollet2017xception}, the decoder modules are trained separately and sometimes in multiple stages. To overcome these drawbacks, \cite{jegou2017one} proposed an equivalent u-net based on the Densenet~\cite{huang2017densely} architecture. However, DenseNet is memory intensive, and adding additional decoder layers leads to a further increase in memory usage. Given these drawbacks, the effectiveness of these architectures on different datasets and applications is unclear.

{\em The goal of this paper is to realize the benefits of u-nets (small size, easy trainability, high performance) for complex natural image segmentation problems.}
Specifically, we propose a new architecture composed of multiple stacks of u-nets. The network executes repeated processing of both top-down as well as bottom-up features and captures long-distance spatial information at multiple resolutions. The network is trained end-to-end on image classification tasks and can be seamlessly applied to semantic segmentation without any additional modules on top (except for replacing the final classifier). 

Our stacked u-net (SUNet) architecture shares some similarity with other related stacked encoder-decoder structures~\cite{fu2017stacked,newell2016stacked}. Fu \etal~\cite{fu2017stacked} uses multiple stacks of de-convolutional networks (maximum of three) on top of a powerful encoder (DenseNet) while \cite{newell2016stacked} applies multiple stacks of u-net modules for human-pose estimation. However, the processing of features inside each u-net module in~\cite{newell2016stacked} differs from ours. ~\cite{newell2016stacked} replaces each convolutional block with a residual module and utilizes nearest-neighbor upsampling for deconvolution. In contrast, SUNets retain the basic u-net structure from~\cite{ronneberger2015u}. Also, SUNet operates without any intermediate supervision and processes features by progressively downsampling while~\cite{newell2016stacked} operates at fix resolution.

\begin{figure*}[!h]
\vspace{-5mm}
\centering
\includegraphics[width=\linewidth]{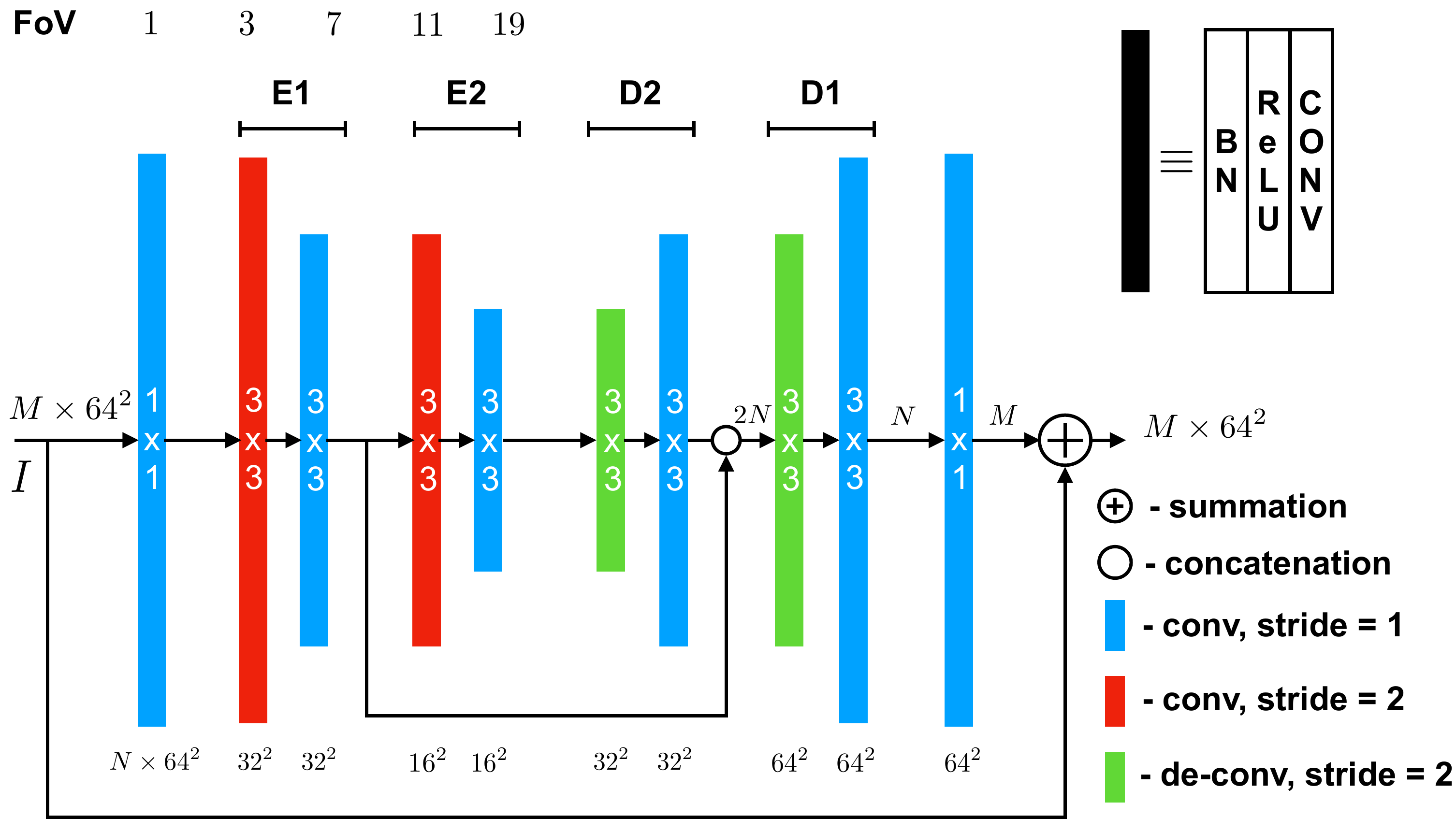}
\caption{A typical u-net module with outer residual connection. $M$ is the number of input features.  Across the u-net module, each layer has the same number of output feature maps (except for the final $1\times 1$ filter), which we denote $N$. For better understanding the figure also includes the field of view (FoV) of each convolutional kernel (top) and the feature map size at the output of each filter (bottom), assuming a $64\times 64$ input $I$. Best viewed in color.}
\label{fig:unet}
\vspace{-10mm}
\end{figure*}

\subsection{U-Net Module Implementation}
Figure \ref{fig:unet} illustrates the design of the u-net module employed in our stacked architecture. Each module is composed of 10 pre-activated convolutional blocks each preceded by a batch-normalization and a ReLU non-linearity. The pooling/unpooling operation, handled by the strided convolutional/deconvolutional layers, facilitates information exchange between the lower and the higher resolution features. A skip connection branches off at the output of the first encoder block, $E1$. Following this, the $E2$ and $D2$ blocks capture long distance context information using lower resolution feature maps and merge the information back with the high resolution features from $E1$ at the output of $D2$. Every layer (except for the bottleneck layers) uses $3\times 3$ kernels, and outputs a fixed number of feature maps, $N$. To mitigate high frequency noise from the sampling operation, each strided conv/de-conv layer is followed by a convolution. Unlike traditional u-nets, the design of convolutional layers in our u-net module helps in retaining the original size of the feature maps at its output (no crop operation takes place). Consequently, multiple u-net modules can be stacked without loosing resolution. 

In the following, we briefly highlight some of the design choices of the architecture. In comparison to traditional u-nets, the max-pooling operation is replaced with strided convolution for SUNets. The use of strided convolutions enables different filters in each u-net module to operate at different resolutions (see the discussion in Section \ref{sec:dsunet}). Moreover, the repeated use of max-pooling operations can cause gridding artifacts in dilated networks~\cite{yu2017dilated}.

Unlike the u-nets of~\cite{ronneberger2015u,newell2016stacked}, our u-net module is comprised of only two levels of depth. We considered two major factors in choosing the depth:  field of view (FoV) of the innermost conv filter and the total number of parameters in a single u-net module. The number of parameters influences the total number of stacks in SUNets. While keeping the total parameters of SUNet approximately constant, we experimented with a higher depth of three and four. We found that the increase in depth indeed led to a decline in performance for the image classification task. This may not be surprising, given that a SUNet with depth of two is able to stack more u-net modules. Moreover, deeper u-net modules make it harder to train the inner-most convolutional layers due to the vanishing gradients problem~\cite{bengio1994learning}. For instance, in our current design, the maximum length of the gradient path is six. The popular classification networks~\cite{he2016deep,huang2017densely} are known to operate primarily on features with $28^2$ and $14^2$ resolution. At this scale, the effective FoV of $19$ is more than sufficient to capture long-distance contextual information. Moreover, the stacking of multiple u-net modules will also serve to increase the effective FoV of higher layers.

SUNets train best when there is sufficient gradient flow to the bottom-most u-net layers.  To avoid vanishing gradients, we include a skip connection~\cite{he2016deep,huang2017densely} around each u-net module. Also, inspired by the design of bottleneck blocks~\cite{he2016deep}, we also include $1\times 1$ convolutional layers. Bottleneck layers restricts the number of input features to a small number ($N$), avoiding parameter inflation. 

When stacking multiple u-nets it makes sense for each u-net module to reuse the raw feature maps from all the preceding u-net modules. Thus we also explored replacing the identity connection with dense connectivity~\cite{huang2017densely}. This new network is memory intensive\footnote{Restricted by the current implementation of deep learning packages} which in turn prevented proper learning of the batch-norm parameters. Instead, we chose to utilize dense connectivity only within each u-net, i.e., while reusing feature maps from $E1$ at $D1$. Thus the proposed u-net module leverages skip connectivity without getting burdened.

\section{SUNets: Stacked U-Nets for Classification}
\begin{table*}[!h]
\centering
\ra{1.05}
\resizebox{1\linewidth}{!}{
\begin{tabular}{@{}c@{\hspace{1mm}}|@{\hspace{1mm}}c@{\hspace{1mm}}|@{\hspace{1mm}}c@{\hspace{1mm}}|@{\hspace{1mm}}c@{\hspace{1mm}}|@{\hspace{1mm}}c@{}}
\toprule
Layers & Output Size & SUNet-64 & SUNet-128 & SUNet-7-128  \\
\midrule
Convolution & $112\times 112$ &  \multicolumn{3}{c}{$7\times 7$ conv, 64, stride 2} \\
\midrule
$\begin{matrix} \text{Residual }\\\text{Block} \end{matrix}$& $56\times 56$ & \multicolumn{3}{c}{$
\begin{bmatrix}
3\times 3 \text{ conv, 128, stride 2} \\
3\times 3 \text{ conv, 128, stride 1}
\end{bmatrix}
\times 1$ }\\
\midrule
$\begin{matrix} \text{UNets Block}\\\text{(1)} \end{matrix}$ & $56\times 56$ & $
\begin{bmatrix}
1\times 1 \text{ conv, 64} \\
\text{U-Net, N=64} \\
1\times 1 \text{ conv, 256}
\end{bmatrix}
\times 2$ 
& $
\begin{bmatrix}
1\times 1 \text{ conv, 128} \\
\text{U-Net, N=128} \\
1\times 1 \text{ conv, 512}
\end{bmatrix}
\times 2$
& $
\begin{bmatrix}
1\times 1 \text{ conv, 128} \\
\text{U-Net, N=128} \\
1\times 1 \text{ conv, 512}
\end{bmatrix}
\times 2$ \\
\midrule
Transition Layer & $28\times 28$ & \multicolumn{3}{c}{$2\times 2$ average pool, stride 2} \\
\midrule
$\begin{matrix} \text{UNets Block}\\\text{(2)} \end{matrix}$ & $28\times 28$ 
& $
\begin{bmatrix}
1\times 1 \text{ conv, 64} \\
\text{U-Net, N=64} \\
1\times 1 \text{ conv, 512}
\end{bmatrix}
\times 4$ 
& $
\begin{bmatrix}
1\times 1 \text{ conv, 128} \\
\text{U-Net, N=128} \\
1\times 1 \text{ conv, 1024}
\end{bmatrix}
\times 4$
& $
\begin{bmatrix}
1\times 1 \text{ conv, 128} \\
\text{U-Net, N=128} \\
1\times 1 \text{ conv, 1280}
\end{bmatrix}
\times 7$ \\
\midrule
Transition Layer & $14\times 14$ & \multicolumn{3}{c}{$2\times 2$ average pool, stride 2} \\
\midrule
$\begin{matrix} \text{UNets Block}\\\text{(3)} \end{matrix}$ & $14\times 14$ 
& $
\begin{bmatrix}
1\times 1 \text{ conv, 64} \\
\text{U-Net, N=64} \\
1\times 1 \text{ conv, 768}
\end{bmatrix}
\times 4$ 
& $
\begin{bmatrix}
1\times 1 \text{ conv, 128} \\
\text{U-Net, N=128} \\
1\times 1 \text{ conv, 1536}
\end{bmatrix}
\times 4$
& $
\begin{bmatrix}
1\times 1 \text{ conv, 128} \\
\text{U-Net, N=128} \\
1\times 1 \text{ conv, 2048}
\end{bmatrix}
\times 7$ \\
\midrule
Transition Layer & $7\times 7$ & \multicolumn{3}{c}{$2\times 2$ average pool, stride 2} \\
\midrule
$\begin{matrix} \text{UNets Block}\\\text{(4)} \end{matrix}$ & $7\times 7$ 
& $
\begin{bmatrix}
1\times 1 \text{ conv, 64} \\
\text{U-Net}^+\text{, N=64} \\
1\times 1 \text{ conv, 1024}
\end{bmatrix}
\times 1$ 
& $
\begin{bmatrix}
1\times 1 \text{ conv, 128} \\
\text{U-Net}^+\text{, N=128} \\
1\times 1 \text{ conv, 2048}
\end{bmatrix}
\times 1$
& $
\begin{bmatrix}
1\times 1 \text{ conv, 128} \\
\text{U-Net}^+\text{, N=128} \\
1\times 1 \text{ conv, 2304}
\end{bmatrix}
\times 1$ \\
\midrule
\multirow{2}{*}{$\begin{matrix} \text{Classification}\\\text{Layer} \end{matrix}$} & $1\times 1$ & \multicolumn{3}{c}{$7\times 7$ global average pool} \\
\cline{2-5}
& & \multicolumn{3}{c}{1000D fully-connected, softmax} \\
\midrule
\multicolumn{2}{c|}{Total Layers} & 110 & 110 & 170 \\
\midrule
\multicolumn{2}{c|}{Params} & $6.9M$ & $24.6M$ & $37.7M$ \\
\bottomrule
\end{tabular}
}
\vspace{.5mm}
\caption{SUNet architectures for ImageNet. $N$ denotes the number of filters per convolutional layer. Note that the building block in bracket refers to the integrated u-net module shown in Figure \ref{fig:unet}.}
\label{tab:SUNET}
\vspace{-10mm}
\end{table*}
Before addressing segmentation, we describe a stacked u-net (SUNet) architecture that is appropriate for image classification.  Because the amount of labeled data available for classification is much larger than for segmentation, classification tasks are often used to pre-train feature extraction networks, which are then adapted to perform segmentation.

The network design of SUNets for ImageNet classification is summarized in Table \ref{tab:SUNET}. Note that each ``conv" layer shown in the table corresponds to a sequence of ``BN-ReLU-Conv" layers. The three listed configurations mainly differ in the number of output feature maps $N$ of each convolutional layer and the total number of stacks in blocks $2$ and $3$. Input images are processed using a $7\times 7$ conv filter followed by a residual block. Inspired by the work on dilated resnet~\cite{yu2017dilated}, the conventional max-pooling layer at this stage is replaced by a strided convolutional layer inside the residual block. Subsequently, the feature maps are processed bottom-up as well as top-down by multiple stacks of u-nets at different scales and with regularly decreasing resolutions. The feature map input size to block $4$ is $7\times 7$ and is further reduced to $2\times 2$ at the input to the encoder $E2$ of the u-net module. At this resolution, it is not possible to have $E2$ and $D2$ layers, and hence a trimmed version of u-nets (u-net$^+$) are employed in block 4. The u-net$^+$ includes a single level of encoder and decoder ($E1,D1$) processing. Towards the end of block 4, a batch normalization is performed and a ReLU non-linearity is applied. Following this, a global average pooling is performed on features and transferred to a softmax classifier.

The residual connection in all but the first u-net in each block is implemented as an identity mapping. In the first u-nets the skip connection is implemented using an expansion layer i.e., a $1\times 1$ conv filter. The number of feature map outputs from each block approximately equates to the total number of feature maps generated by all the preceding u-net modules. This arrangement allows flexibility for the network to retain all the raw feature maps of the preceding modules. Moreover among all other possibilities, the above architectures were picked because their performance on the image classification task is roughly equivalent to the ResNet-18, 50 and 101 network architectures (discussed in Section \ref{sec:exp}), albeit with fewer parameters. However, in contrast to the work on residual net~\cite{zagoruyko2016wide}, our experimentation with wider nets (i.e., $N > 128$) did not yield any performance improvements on ImageNet.

As in ResNet~\cite{he2016deep} and DenseNet~\cite{huang2017densely}, most of the processing in SUNet is performed at the feature scale of $14\times 14$ (46 conv layers) and $7\times 7$ (44 conv layers). However, the order at which the local information is processed can lead to a substantial gap in performance between ResNet and SUNet when extending these popular architectures to object localization, detection, and image segmentation tasks. All these task demands pixel-level localization and hence require a deep architecture that can efficiently integrate local and global cues. The development of SUNet is a first step towards achieving this objective. Intuitively, multiple stacks of u-nets can be seen as multiple iterations of the message passing operation in a CRF~\cite{zheng2015conditional}.

\section{Dilated SUNets for Segmentation}
\label{sec:dsunet}

\begin{figure*}[b]
\vspace{-3mm}
\centering
\includegraphics[width=\linewidth]{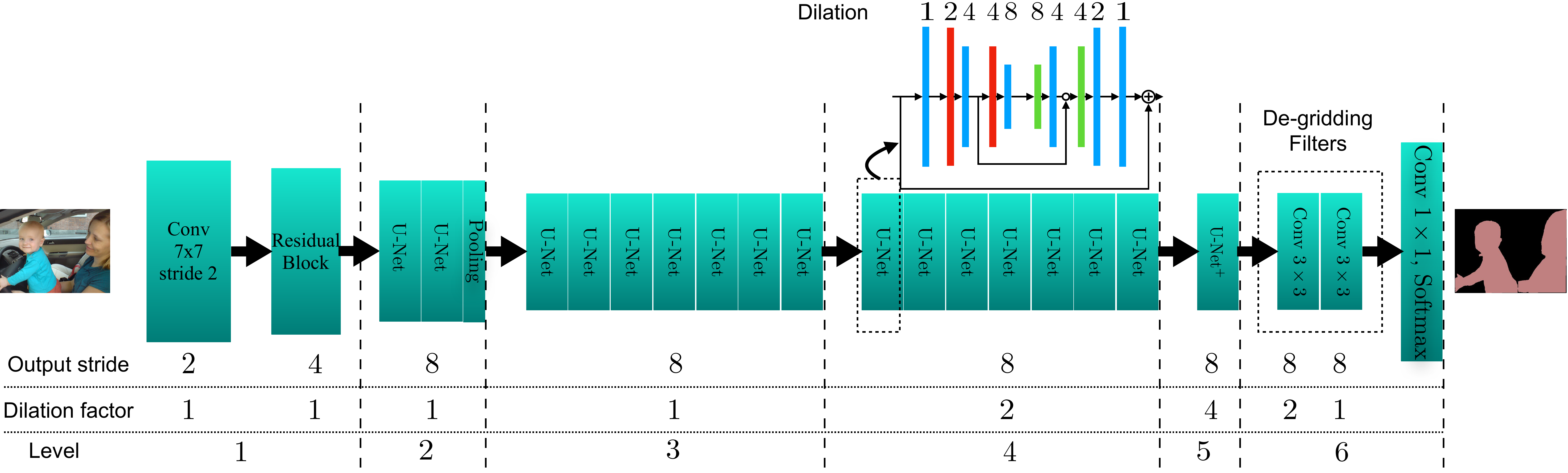}
\caption{Dilated SUNet-7-128 network for segmentation at $output\_stride=8$. For dilation $> 1$, the feature maps are processed with a varying range of dilation factors inside each u-net module (for example, see inset). The de-gridding filters smooth out aliasing artifacts that occur during deconvolution.}
\label{fig:seg}
\vspace{-3mm}
\end{figure*}
We now explain how pre-trained SUNet models can be adapted to perform semantic segmentation (see Section \ref{sec:exp}). One can directly extend SUNet to segmentation by removing a global average pooling layer (to increase output resolution) and operating the network in fully convolutional mode. Akin to other works on semantic segmentation~\cite{zhao2017pyramid,lin2017refinenet,chen2017rethinking}, the output feature maps are rescaled using bilinear interpolation to the input image size before passing into the softmax layer with multi-class cross-entropy loss.

\subsection{Dilation}
For an input image of $512\times 512$, the output map size at the softmax is $16\times 16$ i.e., subsampled by a factor of $32$. This is insufficient to preserve precise pixel-level localization information at its output. The precision can be improved by increasing the output map size of the network. This is realized by dropping the pooling stride at the transition layer. Merely eliminating stride leads to the reduction in the receptive field of the subsequent layers by a factor of two. Consequently, this reduces the influence of long-distance context information on the output prediction. Nevertheless, the receptive field is restored to that of the original network by operating each convolutional filter in the subsequent layers at a dilation factor of 2~\cite{chen2016deeplab,yu2015multi}. 

\subsection{Multigrid}
Figure~\ref{fig:seg} shows a sample dilated SUNet architecture used for the semantic segmentation task. Similar to~\cite{chen2017rethinking}, we define $output\_stride$ to be the ratio of resolution of an input image to that of its output feature map. 

To sample at an $output\_stride$ of $8$ the pooling layers preceding blocks $(3)$ and $(4)$ are discarded. Following this, the dilation factor for each u-net module in blocks $3$ and $4$ is fixed at $2$ and $4,$ respectively. In each subsequent u-net module the $3\times 3$ conv layers are operated with $stride=1$. To keep the receptive field of the low-resolution layers in these modules constant, a dilation is applied. This arrangement facilitates the network to preserve spatial information learned from the prior modules (because there is no downsampling in the final u-net block) while preserving the distance scale of features within each block. As an example, the inset in Figure~\ref{fig:seg} displays the effective dilation rate for each layer in the u-net module at block $3$. Similarly, the dilation rate of each layer (except for bottleneck layers) in the u-net$^+$ module will be twice that of the corresponding layers in block $3$. The steady increase and decrease of dilation factors inside each u-net module is analogous to multigrid solvers for linear systems~\cite{brandt1977multi,briggs2000multigrid}, which use grids at different scales to move information globally. Many recent works~\cite{dai2017deformable,wang2017understanding,chen2017rethinking} on deep networks advocate the use of special structures for information globalization. In SUNet, the multigrid structure is baked into the model, and no further ``frills'' are needed to globalize information.

\subsection{De-gridding Filters}
By adopting dilated SUNets, we observe a vast improvement in segmentation performance. However, for $output\_stride=8$ the segmentation map displays gridding artifacts~\cite{wang2017understanding,yu2017dilated}. This aliasing artifact is introduced when the sampling rate of the dilated layer is lower than the high-frequency content of input feature maps. The final $3\times 3$ conv filter of u-net$^+$ operates at the dilation factor of 4. Directly transferring u-net$^+$'s feature map output to a classification layer can cause gridding artifacts. Following~\cite{yu2017dilated}, the u-net$^+$ module is followed by two layers of de-gridding filters with progressively decreasing dilation factor. Each filter is a $3\times 3$ conv layer and outputs $512$ feature maps.

SUNet does not require any additional post-hoc structural changes popularized by recent works such as decoding layers~\cite{lin2017refinenet,fu2017stacked}, appending context aggregation blocks~\cite{chen2017rethinking,zhao2017pyramid,yu2015multi,chen2016deeplab} and learning conditional random fields~\cite{zheng2015conditional,chandra2017dense}. Hence we regard SUNet as a ``no-frills'' network.

\label{sec:exp}
\section{Experiments}
\subsection{ImageNet Classification}
In this section, we evaluate three SUNet architectures on the ILSVRC-2012 classification dataset, which contains $1.28M$ training images and $50,000$ images for validation, with labels distributed over $1000$ classes. Training utilized the same data augmentation scheme used for ResNet~\cite{he2016deep} and DenseNet~\cite{huang2017densely}. Following common practice~\cite{he2016deep,he2016identity}, we apply a $224\times 224$ center crop on test images and report Top-1 and Top-5 error on the validation set.

{\bf Implementation Details:} All the models were implemented using the PyTorch deep learning framework and trained using four P6000 GPUs on a single node. We use SGD with a batch size of 256. For our largest model, 7-128, we were limited to a batch size of $212,$ due to the GPUs memory constraints. The initial learning rate was set to $0.01$ and decreased by a factor of $10$ every $30$ epochs. We use a weight decay of $5e^{-4}$ and Nesterov momentum of $0.9$ without dampening.  The weights were initialized as in~\cite{he2015delving} and all the models were trained from scratch for a total of $100$ epochs.

Table~\ref{tab:imagenet} compares the performance of SUNet against other classification networks. The comparison is restricted to only ResNet and DenseNet models as most recent work on segmentation builds on top of them. The notable point about the result is that the repeated top-down and the bottom-up processing of features performs equivalently to state-of-the-art classification networks. 

We emphasize that our objective is not to surpass classification accuracy but instead to build a better architecture for segmentation by pre-training on a classification task. Indeed, each SUNet model was selected such that it is the counterpart for the corresponding ResNet model.

\begin{table*}[tb]
\parbox{0.60\linewidth}{
\centering
\ra{1.05}
\resizebox{0.95\linewidth}{!}{
\begin{tabular}{@{}c@{\hspace{1mm}}|@{\hspace{1mm}}c@{\hspace{1mm}}|@{\hspace{1mm}}c@{\hspace{1mm}}|@{\hspace{1mm}}c@{\hspace{1mm}}|@{\hspace{1mm}}c@{}}
\toprule
Model & Top-1 & Top-5 & Depth & Params  \\
\midrule
ResNet-18$^\dagger$ & $30.24$ & $10.92$ & 18 & $11.7M$ \\
ResNet-50$^\dagger$ & $23.85$ & $7.13$ & 50 & $25.6M$\\
ResNet-101$^\dagger$ & $22.63$ & $6.44$ & 101 & $44.5M$\\
\midrule
DenseNet-201$^\dagger$ & $22.80$ & $6.43$ & 201 & $20M$\\
DenseNet-161$^\dagger$ & $22.35$ & $6.20$ & 161 & $28.5M$\\
\midrule
SUNet-64 & $29.28$ & $10.21$ & 111 & $6.9M$ \\
SUNet-128 & $23.64$ & $7.42$ & 111 & $24.6M$ \\
SUNet-7-128 & $22.47$ & $6.85$ & 171 & $37.7M$ \\
\bottomrule
\end{tabular}
}
\vspace{1mm}
\caption{Error rates for classification networks on the ImageNet 2012 validation set. $'\dagger '$ denotes error rates from the official PyTorch implementation.}
\label{tab:imagenet}
}
\hspace{1mm}
\parbox{0.37\linewidth}{
\centering
\ra{1.05}
\resizebox{0.80\linewidth}{!}{
\begin{tabular}{@{}c@{\hspace{1mm}}|@{\hspace{1mm}}c@{}}
\toprule
Model & mIoU \\
\midrule
ResNet-101~\cite{chen2017rethinking} & 68.39 \\
\midrule
SUNet-64 & 72.85 \\
SUNet-128 & 77.16 \\
SUNet-7-128 & 78.95 \\
\bottomrule
\end{tabular}
}
\vspace{1mm}
\caption{The semantic segmentation performance of dilated SUNet and ResNet-101 networks on PASCAL VOC 2012 validation set trained with $output\_stride=16$. Relative to the ResNet-101 network, all SUNets perform very well.}
\label{tab:semseg}
}
\vspace{-8mm}
\end{table*}

\section{Semantic Segmentation}
Semantic segmentation networks were built using the dilated version of the ImageNet pre-trained SUNet models (Section \ref{sec:dsunet}). We evaluate on the PASCAL VOC 2012 semantic segmentation benchmark~\cite{everingham2015pascal} and urban scene understanding Cityscape~\cite{cordts2016cityscapes} datasets. The performance on each of these datasets is reported using intersection-over-union (IoU) averaged over all classes.

\subsection{Datasets}

{\bf PASCAL VOC 2012:} This dataset contains $1{,}464$ train, $1{,}449$ validation and $1{,}456$ test images. The pixel-level annotation for $20$ objects and one background class is made available for the train and validation set. Following common practice, the train set is augmented with additional annotated data from~\cite{BharathICCV2011} which finally provides a total of $10{,}582$ (trainaug) training images.

\noindent
{\bf Cityscape:} This dataset consists of finely annotated images of urban scenes covering multiple instances of cars, roads, pedestrians, buildings, etc. In total, it contains $19$ classes on $2,975$ finely annotated training and $500$ validation images.

\subsection{Implementation Details}
We use the SGD optimizer with a momentum of $0.95$ and weight decay of $10^{-4}$. Each model is fine-tuned starting with an initial learning rate of $0.0002$ which is decreased every iteration by a factor of $0.5 \times \left(1+ \cos \left(\pi  \frac{\text{iter}}{\text{max iters}}\right)\right)$~\cite{loshchilov2016sgdr}. The batch-norm parameters are learned with a decay rate of $0.99$ and the input crop size for each training image is set to $512\times 512$. 
We train each model using two P6000 GPUs and the batch size $22$. On PASCAL VOC, each model is trained for $45$K iterations while for Cityscapes we use $90$K iterations.

Unless mentioned, for all our experiments we set $output\_stride=16$. This means only the u-net modules in the final block $(4)$ operate at dilation factor of two; all other modules use the same stride as in the original classification model. Furthermore, $output\_stride=16$ enables larger batch sizes than smaller stride choices, and hence leads to efficient learning of batch norm parameters.
Also, training and inference are $2\times$ faster with $output\_stride=16$ rather than ${8}$. 

\noindent
{\bf Data Augmentation:} To prevent overfitting during the training process, each training image is resized with a random scale from $0.5$ to $2$ following which the input image is randomly cropped. Additionally, the input is randomly flipped horizontally and also randomly rotated between $-10^\circ$ to $10^\circ$.

\subsection{Ablation Study}
We experiment with different SUNet variants on the PASCAL VOC 2012 dataset. 

\noindent
{\bf SUNets vs ResNet-101:} We compare the performance of the dilated SUNet architecture on semantic segmentation against the popular dilated ResNet-101 model. Models were fine-tuned on the ``trainaug'' set without the degridding layers and evaluated on the validation set. 

The performance of the plain dilated SUNets surpasses that of ResNet-101 by a wide margin
when trained with $output\_stride=16$ (Table~\ref{tab:semseg}). In fact, the smallest SUNet model, SUNet-64 with $6.7M$ parameters, beats ResNet-101 (with $44.5M$) by an absolute margin of $4.5\%$ IoU while SUNet-7-128, the counterpart network to ResNet-101, improves by over $10.5\%$ IoU. This is substantial, given that the gap between the ResNet-101 and VGG-16 models is $\sim 3\%$~\cite{chen2016deeplab} (at $output\_stride=8$). This contrasts with the negligible performance differences observed on classification, and suggests that specialized segmentation network architectures can surpass architectures adapted from classification.

Finally, we note that, although SUNets were designed for pixel-level localization tasks, the selected models were chosen only based on their classification performance. By linking the model selection process to the primary task (segmentation and object detection) there is a possibility of improving performance.

\begin{table*}[tb]
\parbox{0.37\linewidth}{
\centering
\ra{1.05}
\resizebox{1\linewidth}{!}{
\begin{tabular}{@{}c@{\hspace{1mm}}|@{\hspace{1mm}}c@{\hspace{1mm}}|@{\hspace{1mm}}c@{}}
\toprule
OS &  Strided conv & Multigrid\\
\midrule
8 & 65.99 & 78.64 \\
16 & 78.25 & 78.95  \\
\bottomrule
\end{tabular}
}
\vspace{.25mm}
\caption{Performance comparison of multigrid dilation against strided convolution inside each u-net module, using the SUNet-7-128 model and evaluated using mean IoU. {\bf OS} denotes $output\_stride$ during training.}
\label{tab:nodilation}
}
\hspace{1mm}
\parbox{0.60\linewidth}{
\centering
\ra{1.05}
\resizebox{0.85\linewidth}{!}{
\begin{tabular}{@{}c@{\hspace{1mm}}|@{\hspace{1mm}}c@{\hspace{1mm}}|@{\hspace{1mm}}c@{\hspace{1mm}}|@{\hspace{1mm}}c@{\hspace{1mm}}|@{\hspace{1mm}}c@{\hspace{1mm}}|@{\hspace{1mm}}c@{}}
\toprule
train OS & eval OS & DL & MS & Flip & mIoU \\
\midrule
32 & 32 & & & & 76.03 \\
32 & 32 & & \checkmark & & 77.58 \\
32 & 32 & & \checkmark & \checkmark & 77.57 \\
\midrule
16 & 16 & & & & 78.95 \\
16 & 16 & \checkmark & & & 78.10 \\
16 & 16 & & \checkmark & & 80.22 \\
16 & 16 & & \checkmark & \checkmark & 80.40 \\
\midrule
8 & 8 & & & & 78.64 \\
8 & 8 & \checkmark & & & 78.88 \\
8 & 8 & & \checkmark & & 80.37 \\
8 & 8 & & \checkmark & \checkmark & 80.50 \\
\bottomrule
\end{tabular}
}
\vspace{1mm}
\caption{Performance comparison at various $output\_stride$ and inference strategies. {\bf MS:} Multi-scale, {\bf DL:} with Degridding Layers}
\label{tab:outputstride}
}
\vspace{-10mm}
\end{table*}

\noindent
{\bf Multigrid vs Downsampling:}
We compare the performance of multigrid dilation (as shown in Figure \ref{fig:seg}) inside each u-net against the usual downsampling (Figure \ref{fig:unet}). We consider a dilated SUNet-7-128 network and report performance at different training $output\_stride$. For $output\_stride=8$, the network was trained with a batch size of $12$. The result is summarized in Table \ref{tab:nodilation}. For a dilated network, replacing strided convolutional layers with the corresponding dilated layers is more logical as well as beneficial. This is because, when operating dilated convolutional layers with $stride > 1$, alternate feature points are dropped without being processed by any of the filters, leading to high frequency noise at the decoder output. Furthermore, due to a skip connection, the features from the lower layers are also corrupted. Due to error propagation, this effect is more prominent in a network with many dilated modules (for eg., $output\_stride=8$). 

\noindent
{\bf Output Stride and Inference Strategy:} Finally, we experiment with three different training output strides (8,16,32) and multi-scale inference at test time. For $output\_stride=32$, none of the layers are dilated and hence de-gridding layers were not used. Training with an $output\_stride=32$ is equivalent to fine-tuning a classification network with a global  pooling layer removed. For multi-scale inference, each input image is scaled and tested using multiple scales $\{0.5, 0.75, 1.0, 1.25\}$  and its left-right flipped image. The average over all output maps is used in the final prediction. See results in Table~\ref{tab:outputstride}. We note that:
\begin{enumerate}
\item The network trained with $OS=32$ performs $0.7$ IoU better (with single scale) than the Resnet-101 and Resnet-152 models~\cite{wu2016wider} each trained at $OS=8$. This is significant, since the SUNet output contains $16\times$ fewer pixels. This leads to $4\times$ faster training/inference without a performance drop.
\item The degridding layers do not improve performance at $OS=16$. In this case, there is only a small change in dilation factor between the final layer of SUNet and the classification layer, so aliasing is not problematic.
\item The margin of performance improvement decreases with increase in training $OS$. Given this and the above fact, subsequently we only report performance for models trained at $OS=16$ without any degridding layers.
\end{enumerate}

\noindent
{\bf Pretraining with MS-COCO:} Following common practice~\cite{zhao2017pyramid,chen2017rethinking,deeplabv3xception}, we pretrain SUNet-7-128 with the MS-COCO dataset~\cite{lin2014microsoft}. The MS-COCO dataset contains pixel-level annotation for $80$ object classes. Except for the PASCAL VOC classes, the pixel annotation for all other classes is set to the background class. Following this, we use all the images from the MS-COCO ``trainval'' set except for those having $<1000$ annotated pixel labels. This yields $90,000$ training images. After pretraining, the model is fine-tuned on ``trainaug'' for $5$K iterations with $10\times$ smaller initial learning rate. In the end, our model achieves $83.27\%$ mIoU on the validation set. This performance is slightly better than the ResNet+ASPP model~\cite{chen2017rethinking} (82.70\%) and equivalent to Xception+ASPP+Decoder model~\cite{deeplabv3xception} (83.34\%).

\subsection{Results on Test set}
\begin{figure}[!h]
\centering
 \begin{tabular}{c c |c c |c c}
     \includegraphics[width=0.16\linewidth]{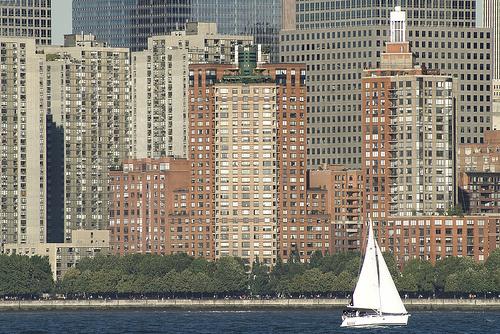}&
     \includegraphics[width=0.16\linewidth]{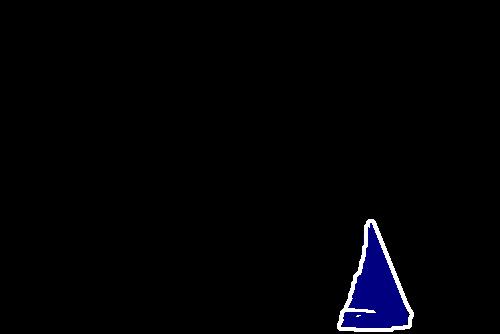}&
     \includegraphics[width=0.16\linewidth]{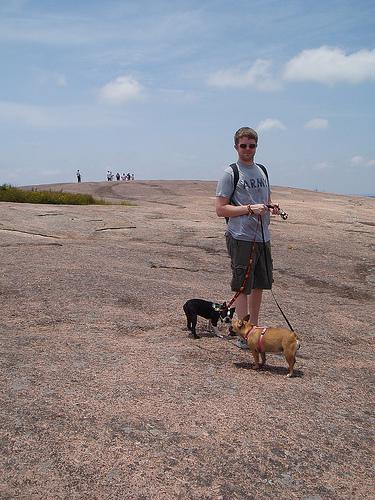}&
     \includegraphics[width=0.16\linewidth]{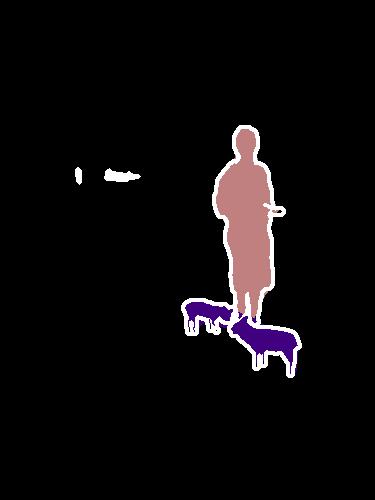}&
     \includegraphics[width=0.16\linewidth]{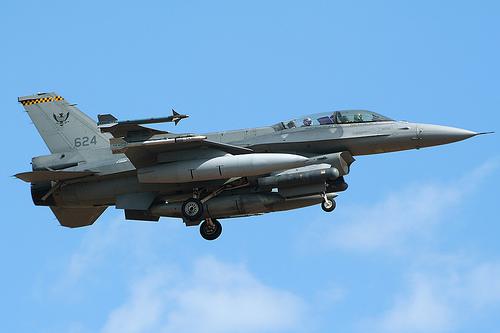}&
     \includegraphics[width=0.16\linewidth]{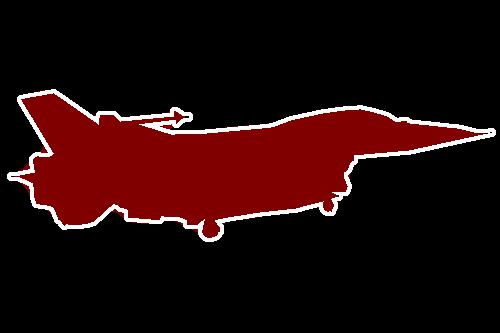}\\[-1ex]
     \includegraphics[width=0.16\linewidth]{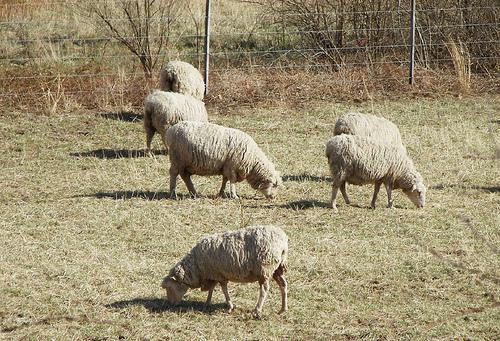}&
     \includegraphics[width=0.16\linewidth]{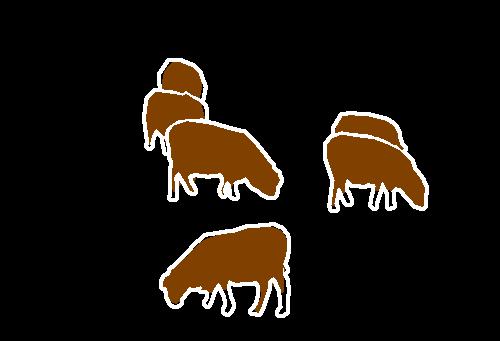}&
     \includegraphics[width=0.16\linewidth]{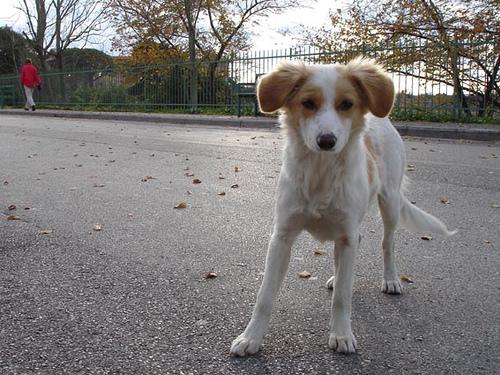}&
     \includegraphics[width=0.16\linewidth]{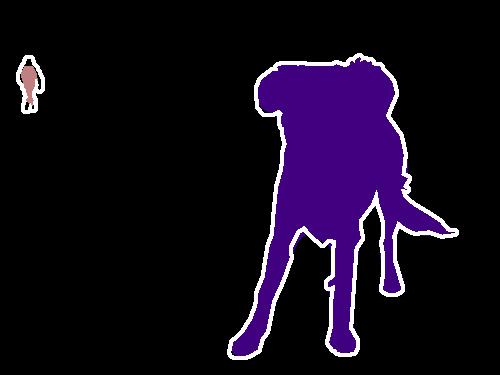}&
     \includegraphics[width=0.16\linewidth]{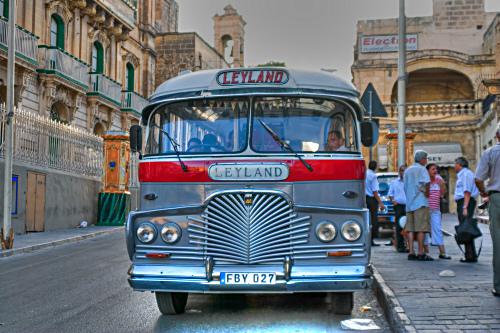}&
     \includegraphics[width=0.16\linewidth]{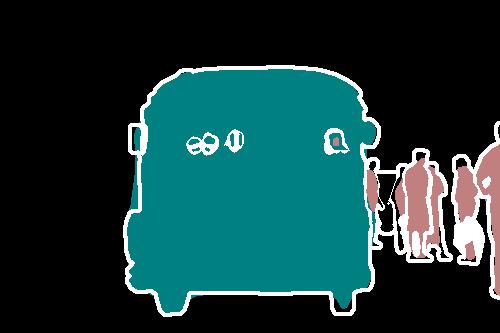}\\
          \hline
   \end{tabular}
    \begin{tabular}{c c c | c c c}
     input & target & output & input & target & output \\
     \includegraphics[width=0.16\linewidth]{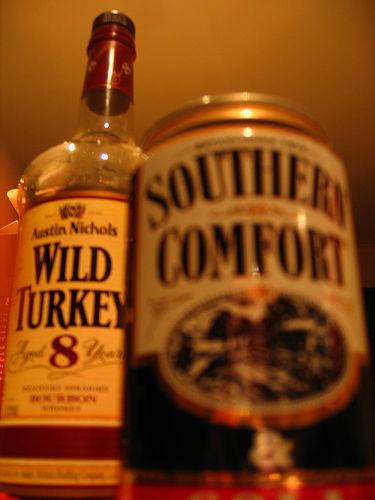}&
     \includegraphics[width=0.16\linewidth]{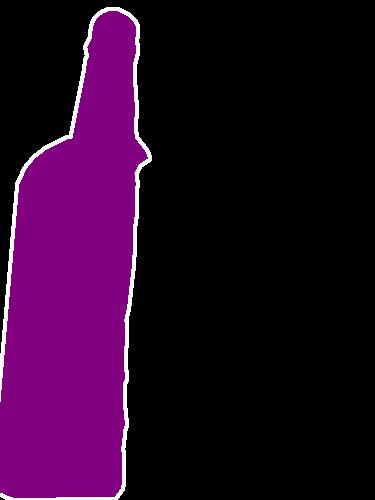}&
     \includegraphics[width=0.16\linewidth]{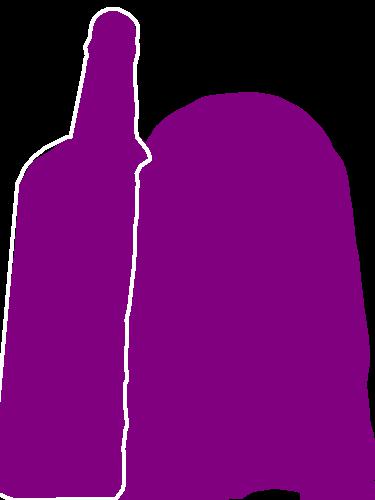}&     
	\includegraphics[width=0.16\linewidth]{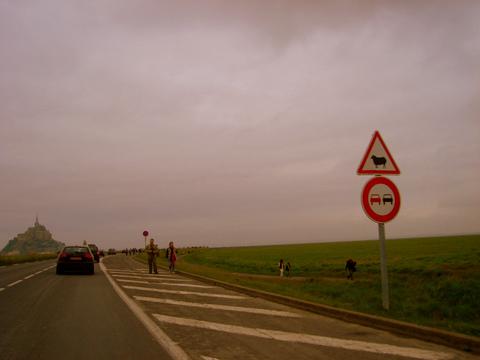}&
     \includegraphics[width=0.16\linewidth]{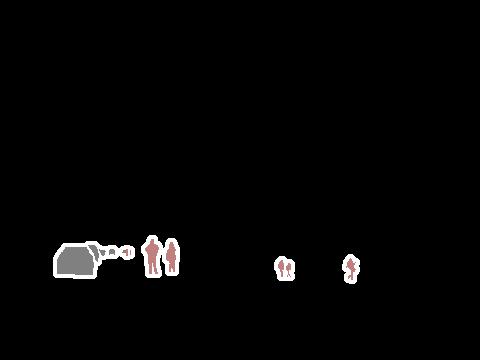}&
     \includegraphics[width=0.16\linewidth]{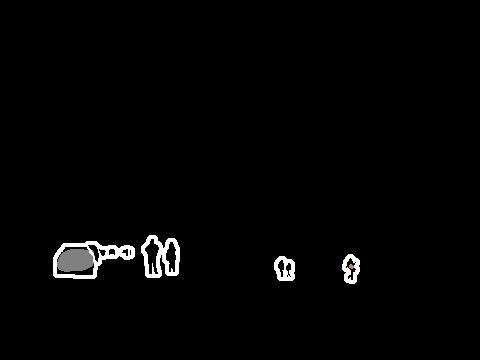}\\
   \end{tabular}
\caption{Visualization of the segmentation output on PASCAL VOC 2012 $val$ set when trained at an $output\_stride=16$ using SUNet-7-128 network + MS-COCO. Final row shows couple of failure case which happens due to, ambiguous annotation and inability in detecting low resolution objects.}
\label{fig:visual}
\end{figure}

\begin{figure}[!h]
\centering
 \begin{tabular}{c c |c c |c c}
     \includegraphics[width=0.16\linewidth]{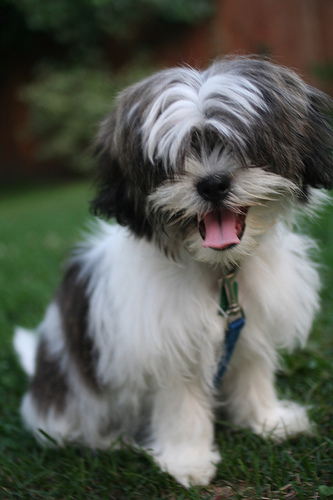}&
     \includegraphics[width=0.16\linewidth]{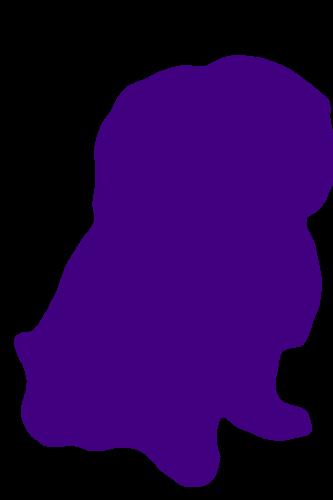}&
          \includegraphics[width=0.16\linewidth]{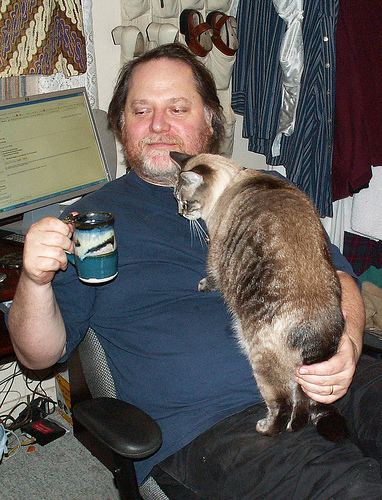}&
     \includegraphics[width=0.16\linewidth]{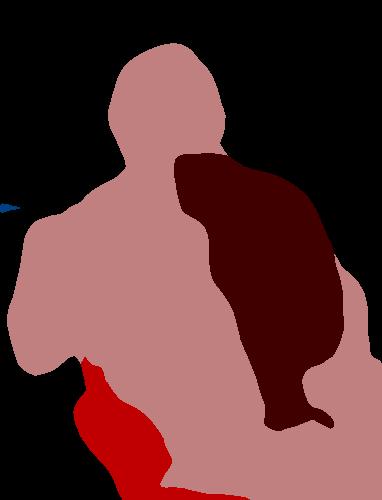}&
          \includegraphics[width=0.16\linewidth]{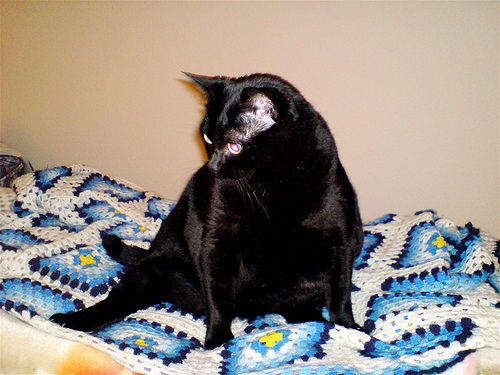}&
     \includegraphics[width=0.16\linewidth]{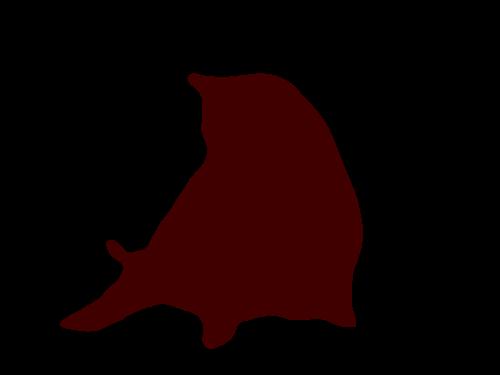}\\
     \includegraphics[width=0.16\linewidth]{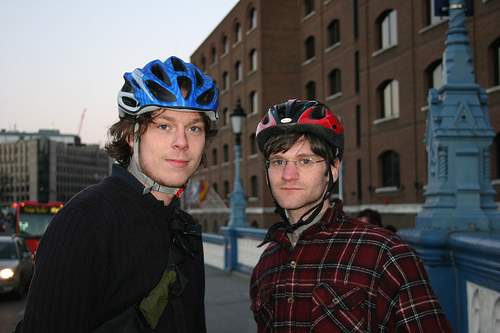}&
     \includegraphics[width=0.16\linewidth]{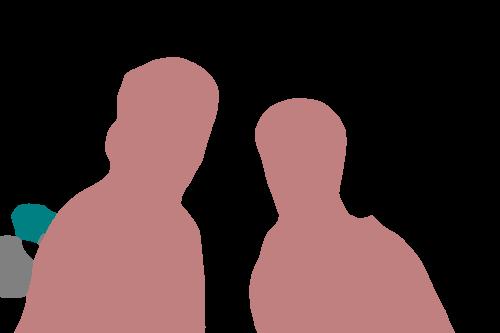}&
     \includegraphics[width=0.16\linewidth]{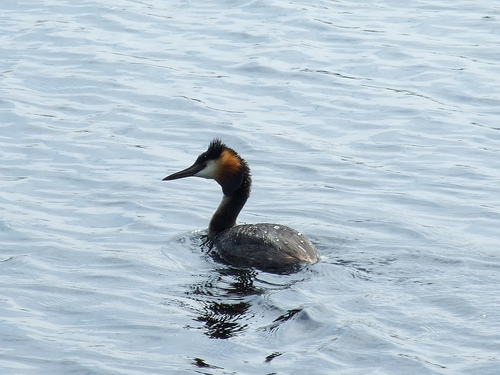}&
     \includegraphics[width=0.16\linewidth]{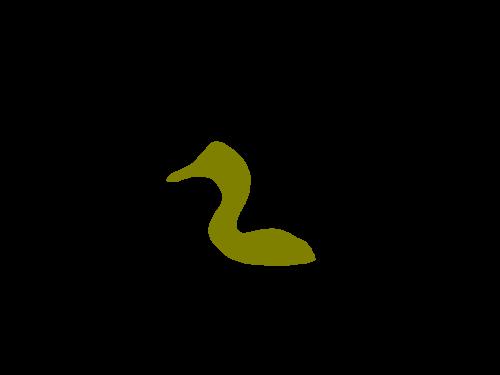}&
     \includegraphics[width=0.16\linewidth]{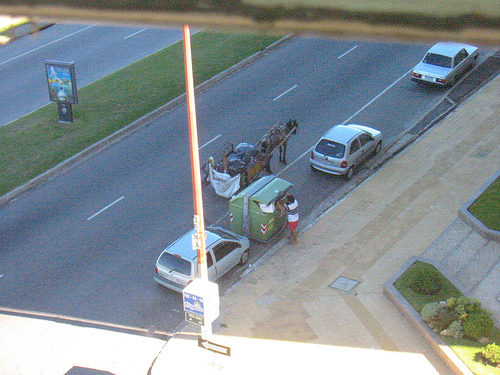}&
     \includegraphics[width=0.16\linewidth]{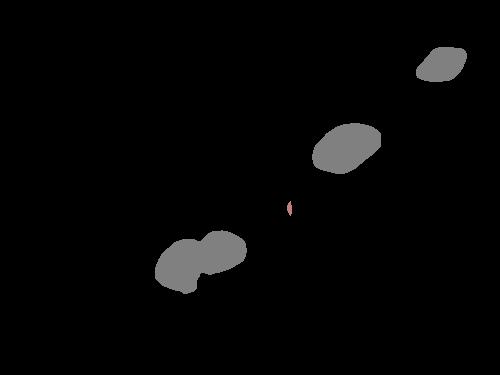}\\
     \includegraphics[width=0.16\linewidth]{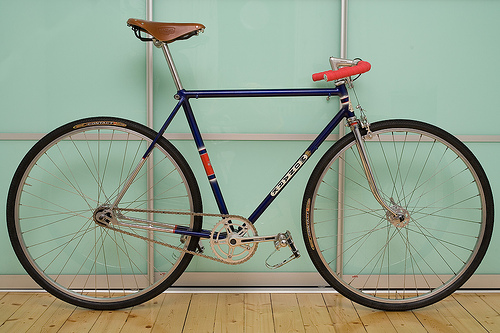}&
     \includegraphics[width=0.16\linewidth]{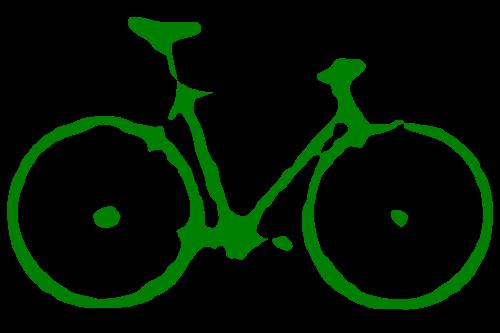}&
     \includegraphics[width=0.16\linewidth]{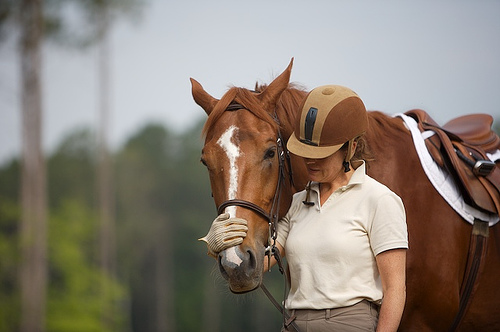}&
     \includegraphics[width=0.16\linewidth]{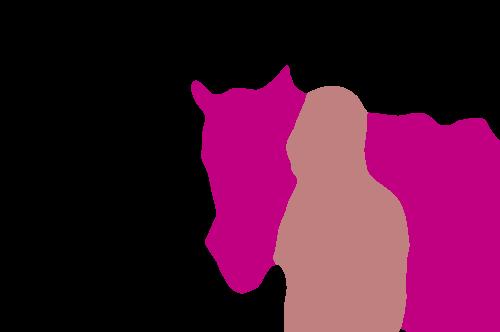}&
     \includegraphics[width=0.16\linewidth]{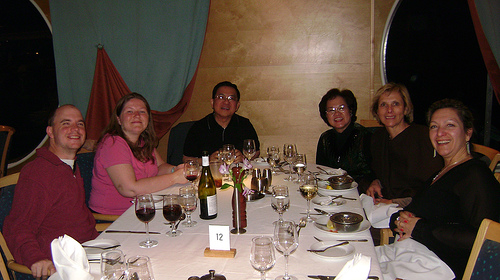}&
     \includegraphics[width=0.16\linewidth]{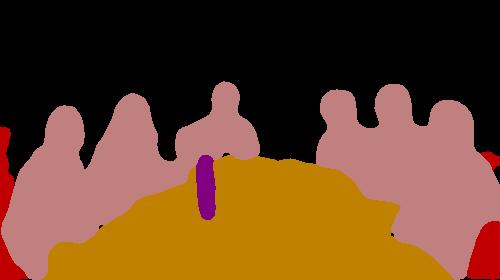}\\
     \includegraphics[width=0.16\linewidth]{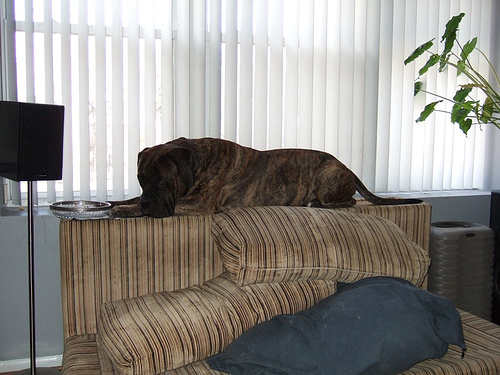}&
     \includegraphics[width=0.16\linewidth]{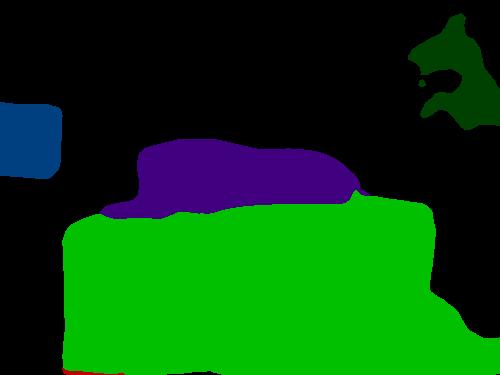}&
     \includegraphics[width=0.16\linewidth]{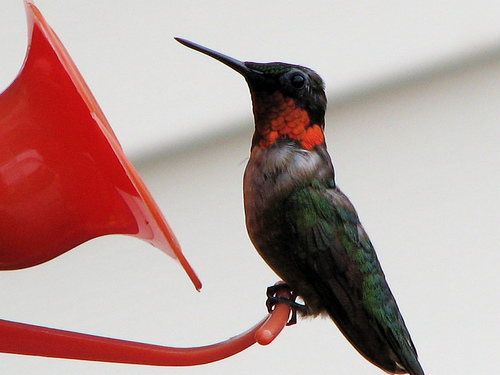}&
     \includegraphics[width=0.16\linewidth]{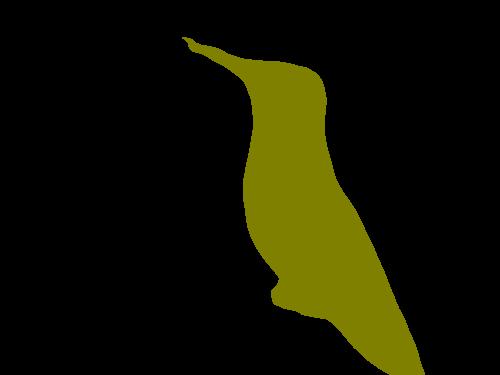}&
     \includegraphics[width=0.16\linewidth]{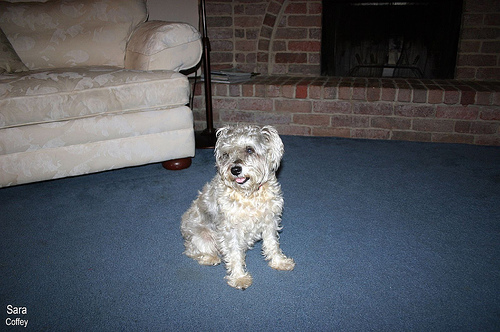}&
     \includegraphics[width=0.16\linewidth]{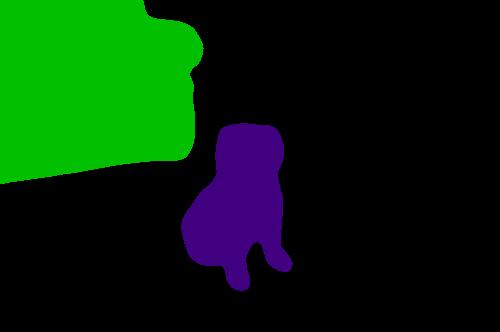}\\
   \end{tabular}
\caption{Visualization of the segmentation output on PASCAL VOC 2012 $test$ set.}
\label{fig:visualtest}
\end{figure}

\noindent
{\bf PASCAL VOC 2012:}  Before submitting test set output to an evaluation server, the above model was further fine-tuned on the ``trainval'' set with batch-norm parameters frozen and at $10\times$ smaller initial learning rate. Table~\ref{tab:pascal} compares the test set results against other state-of-the-art methods.  PSPNet performs slightly better than SUNet, but at the cost of $30M$ more parameters while training at an $output\_stride=8$. Figure~\ref{fig:visual}€ and~\ref{fig:visualtest}€ displays some qualitative results on validation and test sets.

\noindent
{\bf Cityscapes:} A similar training strategy as in PASCAL is adopted except that the multi-scale inference is performed on additional scales $\{1.5,1.75,2.0,2.25,2.5\}$. Only the \textit{finer} annotation set was used for training. The comparison on the Cityscapes test set results are displayed in Table~\ref{tab:cityscape}.

\begin{table}[!tbp]
\parbox{0.48\linewidth}{
\centering
\ra{1.05}
\resizebox{0.9\linewidth}{!}{
\begin{tabular}{@{}c@{\hspace{1mm}}|@{\hspace{1mm}}c@{}}
\toprule
Methods & mIoU \\
\midrule
Piecewise (VGG16) ~\cite{lin2016efficient} & 78.0 \\
LRR+CRF~\cite{ghiasi2016laplacian} & 77.3 \\
DeepLabv2+CRF~\cite{chen2016deeplab} & 79.7 \\
Large-Kernel+CRF~\cite{peng2017large} & 82.2 \\
Deep Layer Cascade$^*$~\cite{li2017not} & 82.7 \\
Understanding Conv~\cite{wang2017understanding} & 83.1 \\
RefineNet~\cite{lin2017refinenet} & 82.4 \\
RefineNet-ResNet152~\cite{lin2017refinenet} & 83.4 \\
\midrule
PSPNet~\cite{zhao2017pyramid} & \textbf{85.4} \\
SUNet-7-128 & 84.3\footnotemark \\
\bottomrule
\end{tabular}
}
\vspace{1mm}
\caption{Performance comparison on PASCAL VOC 2012 test set. For fair comparison, only the methods pre-trained using MS-COCO are displayed.}
\label{tab:pascal}
}
\hspace{1mm}
\parbox{0.48\linewidth}{
\centering
\ra{1.05}
\resizebox{0.9\linewidth}{!}{
\begin{tabular}{@{}c@{\hspace{1mm}}|@{\hspace{1mm}}c@{}}
\toprule
Methods & mIoU \\
\midrule
LRR (VGG16)~\cite{ghiasi2016laplacian} & 69.7 \\
DeepLabv2+CRF~\cite{chen2016deeplab} & 70.4 \\
Deep Layer Cascade$^*$~\cite{li2017not} & 71.1 \\
Piecewise (VGG16)~\cite{lin2016efficient} & 71.6 \\
RefineNet~\cite{lin2017refinenet} & 73.6 \\
\midrule
Understanding Conv~\cite{wang2017understanding} & 77.6 \\
PSPNet~\cite{zhao2017pyramid} & \textbf{78.4} \\
SUNet-7-128 & 75.3\footnotemark \\
\bottomrule
\end{tabular}
}
\vspace{1mm}
\caption{Performance comparison on Cityscapes $test$ set. All methods were trained only using the ``fine" set. All nets utilize ResNet-101 as a base network, except if specified or marked with $^*$.}
\label{tab:cityscape}
}
\vspace{-8mm}
\end{table}
\footnotetext{\url{http://host.robots.ox.ac.uk:8080/anonymous/KT7JGJ.html}}
\footnotetext{\url{https://www.cityscapes-dataset.com/method-details/?submissionID=1151}}

\subsection{Activation Maps}
\begin{figure}[!h]
\centering
 \begin{tabular}{c c c c c }
 $OS=8$ &   $OS=8$ + DL &  $OS=8$+Strided &  $OS=16$ & SUNet-64\\
     \includegraphics[width=0.2\linewidth]{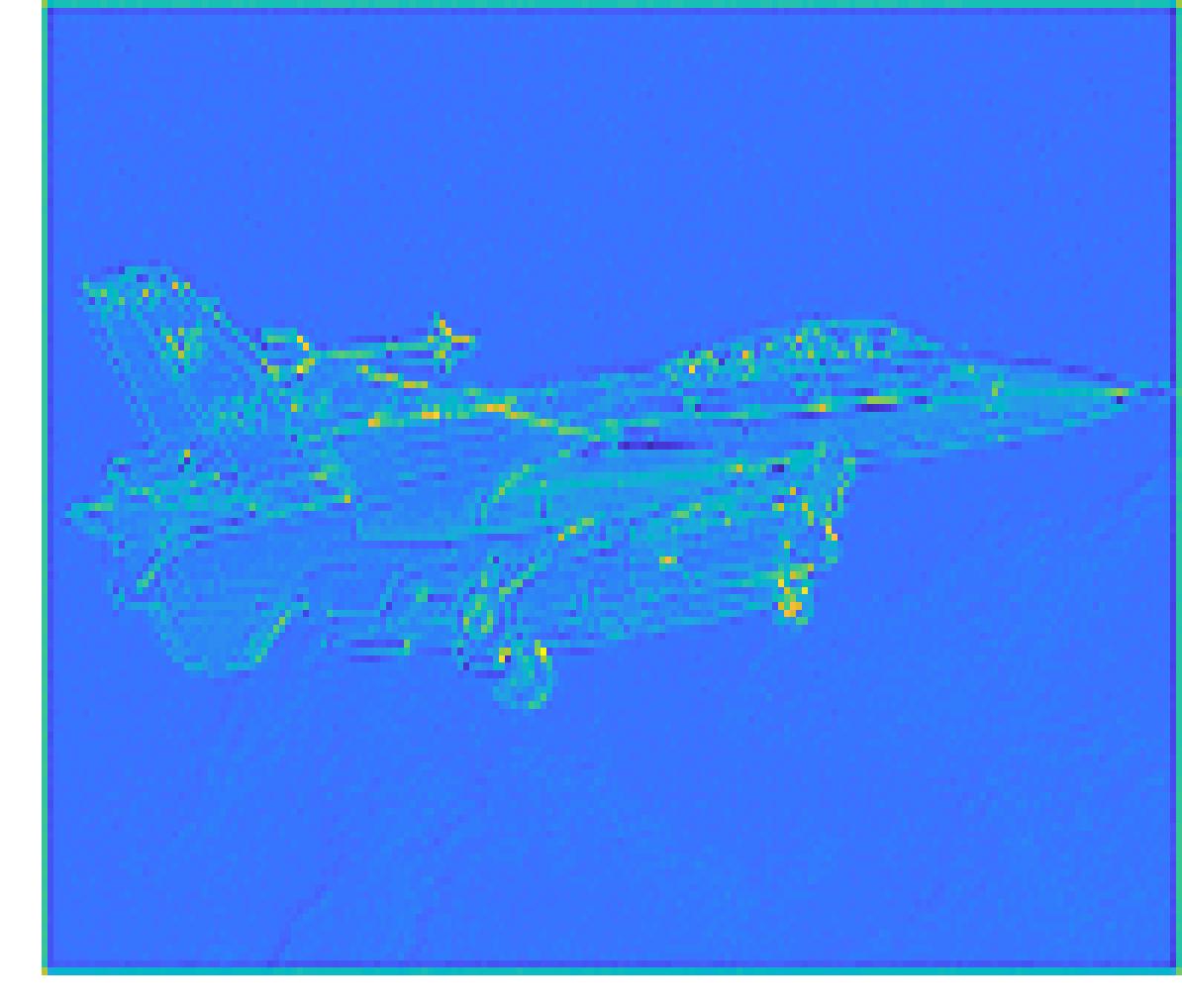}&
     \includegraphics[width=0.2\linewidth]{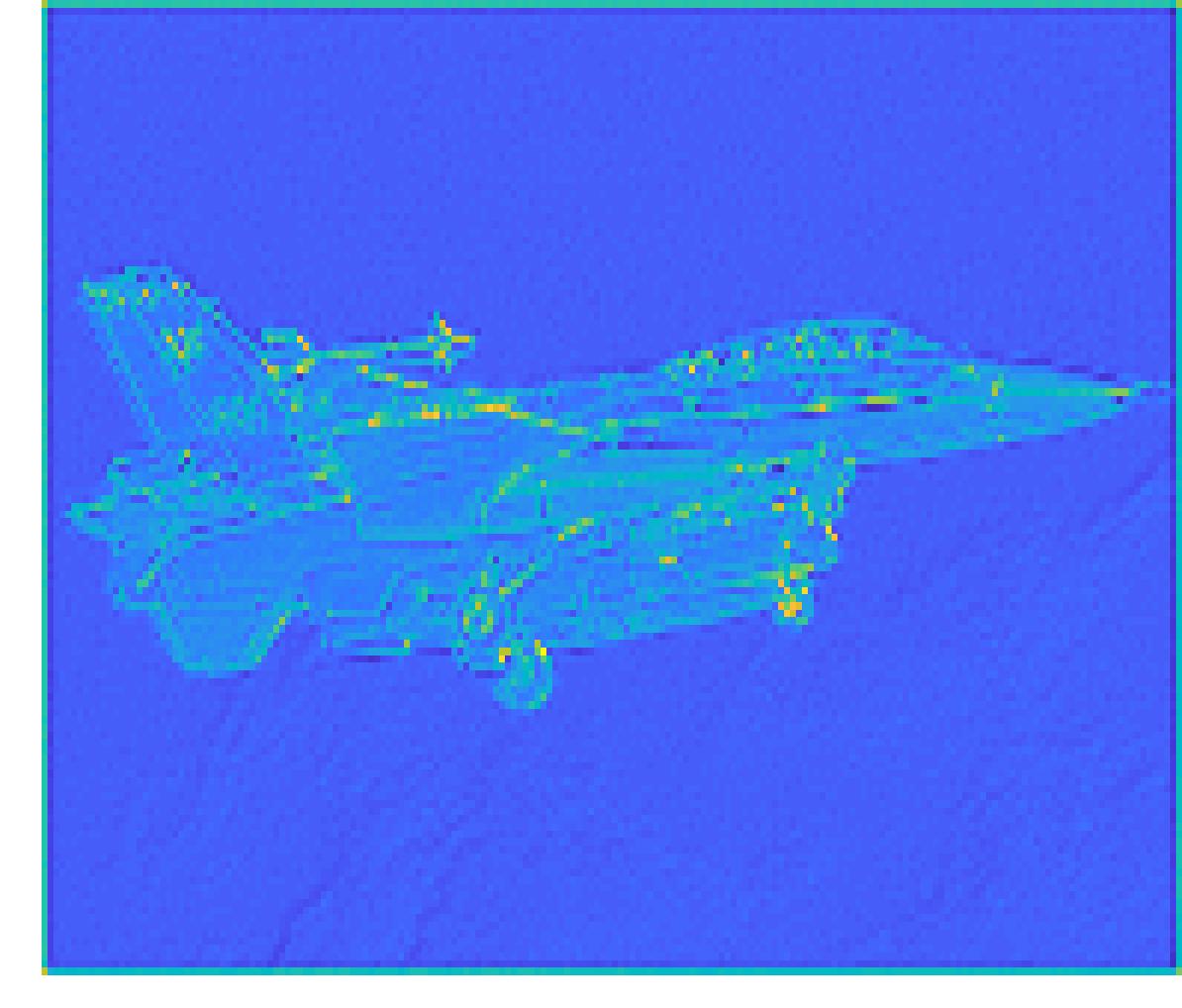}&
     \includegraphics[width=0.2\linewidth]{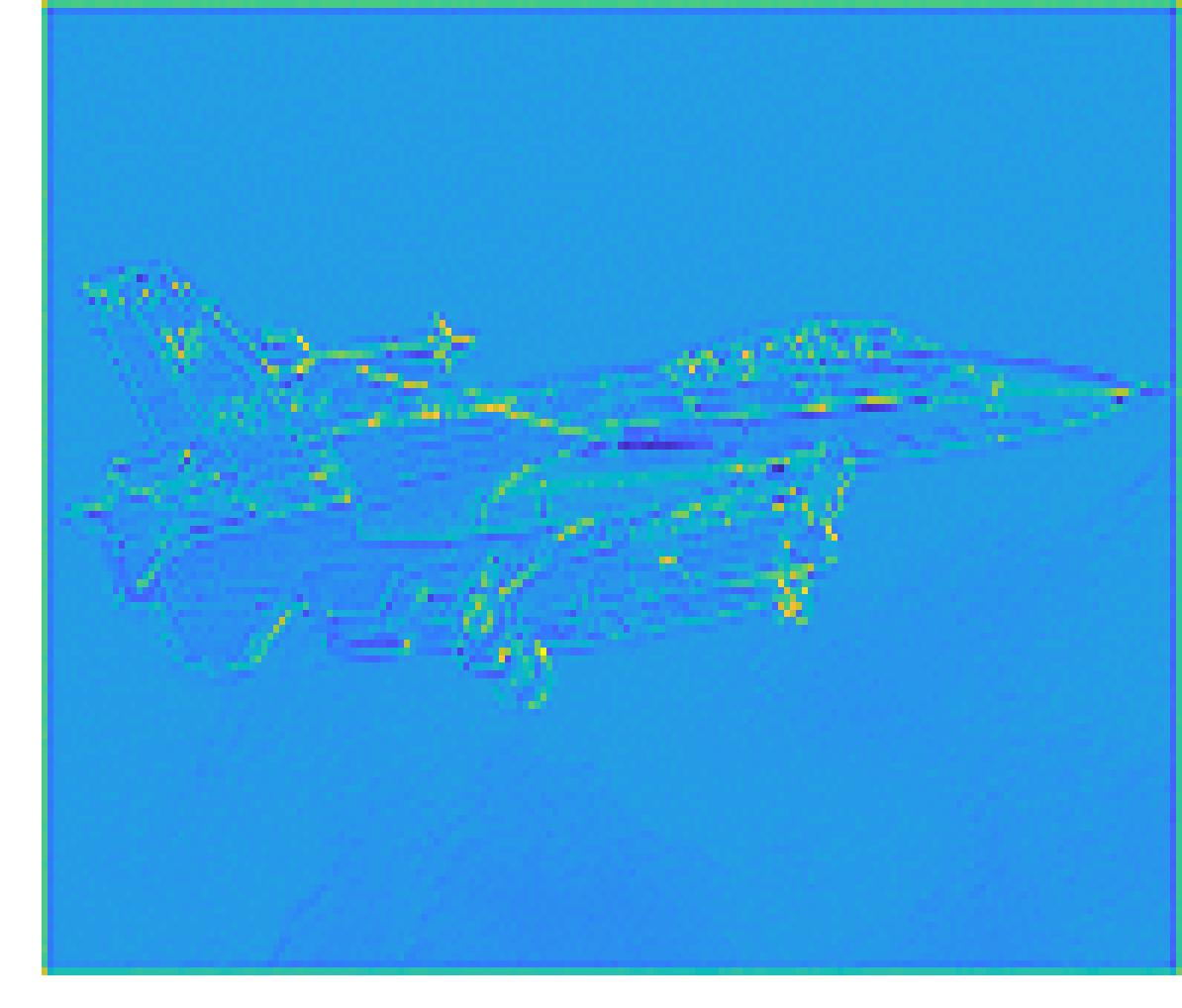}&
     \includegraphics[width=0.2\linewidth]{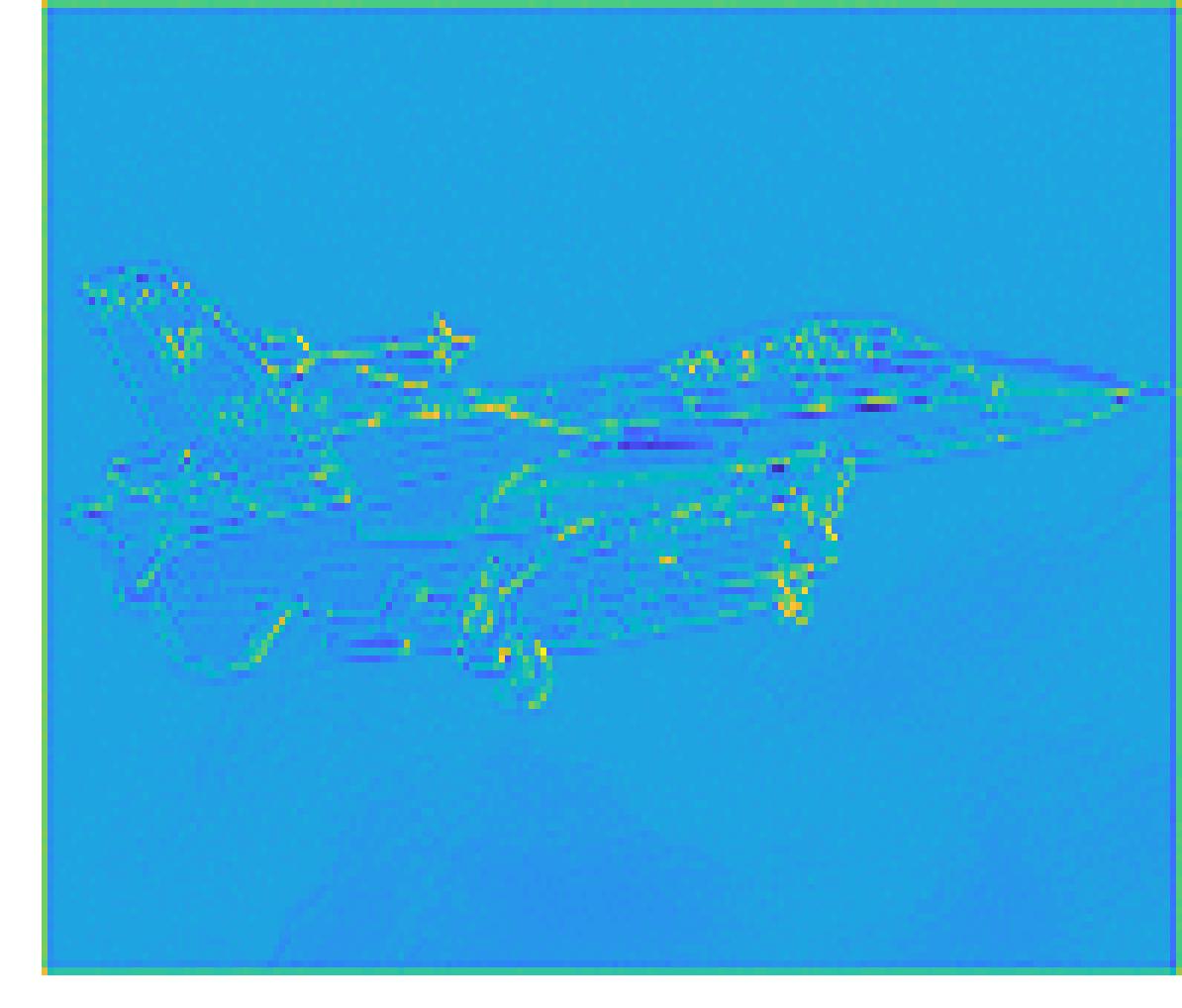}&
     \includegraphics[width=0.2\linewidth]{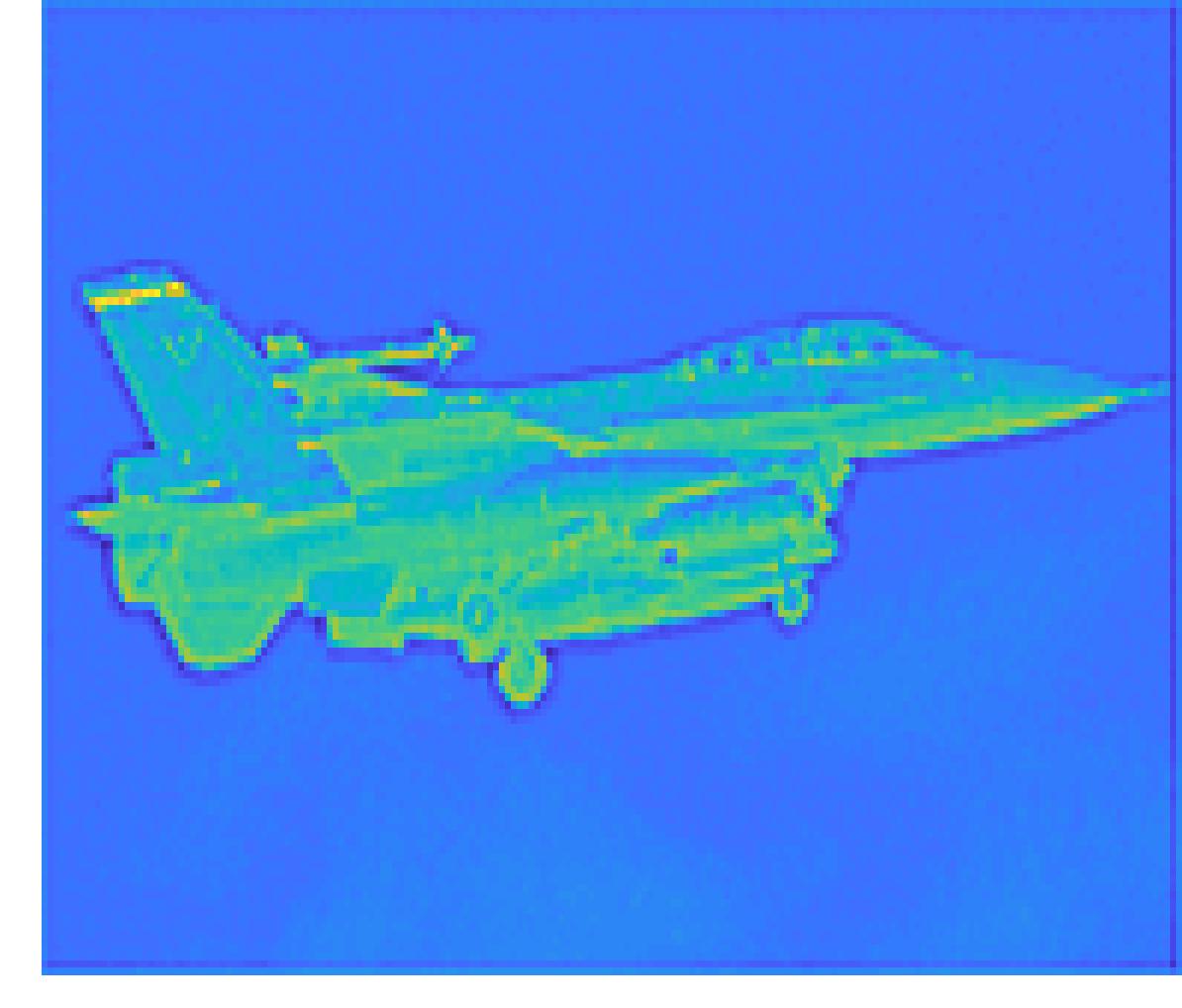}\\[-1ex]
     \includegraphics[width=0.2\linewidth]{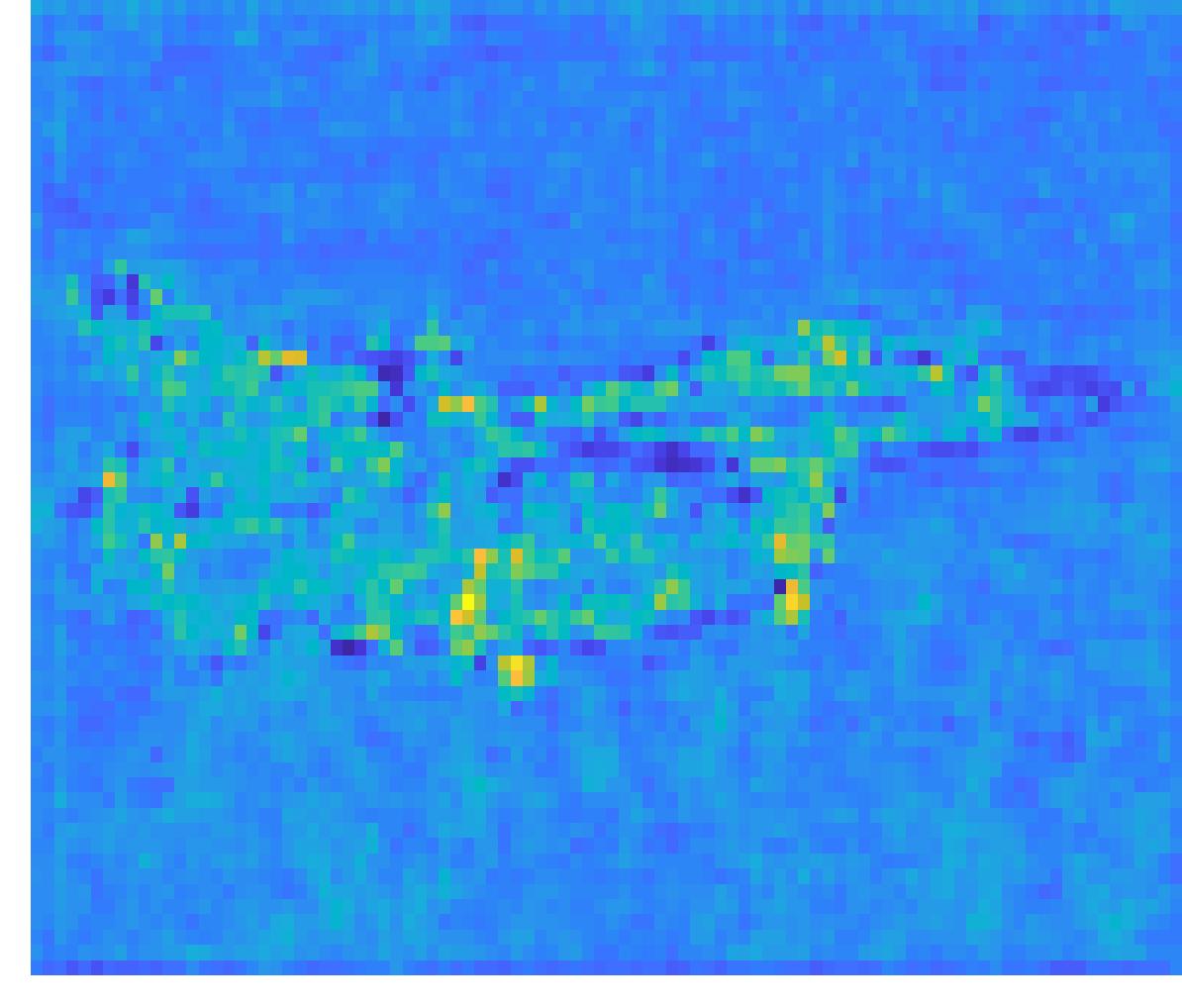}&
     \includegraphics[width=0.2\linewidth]{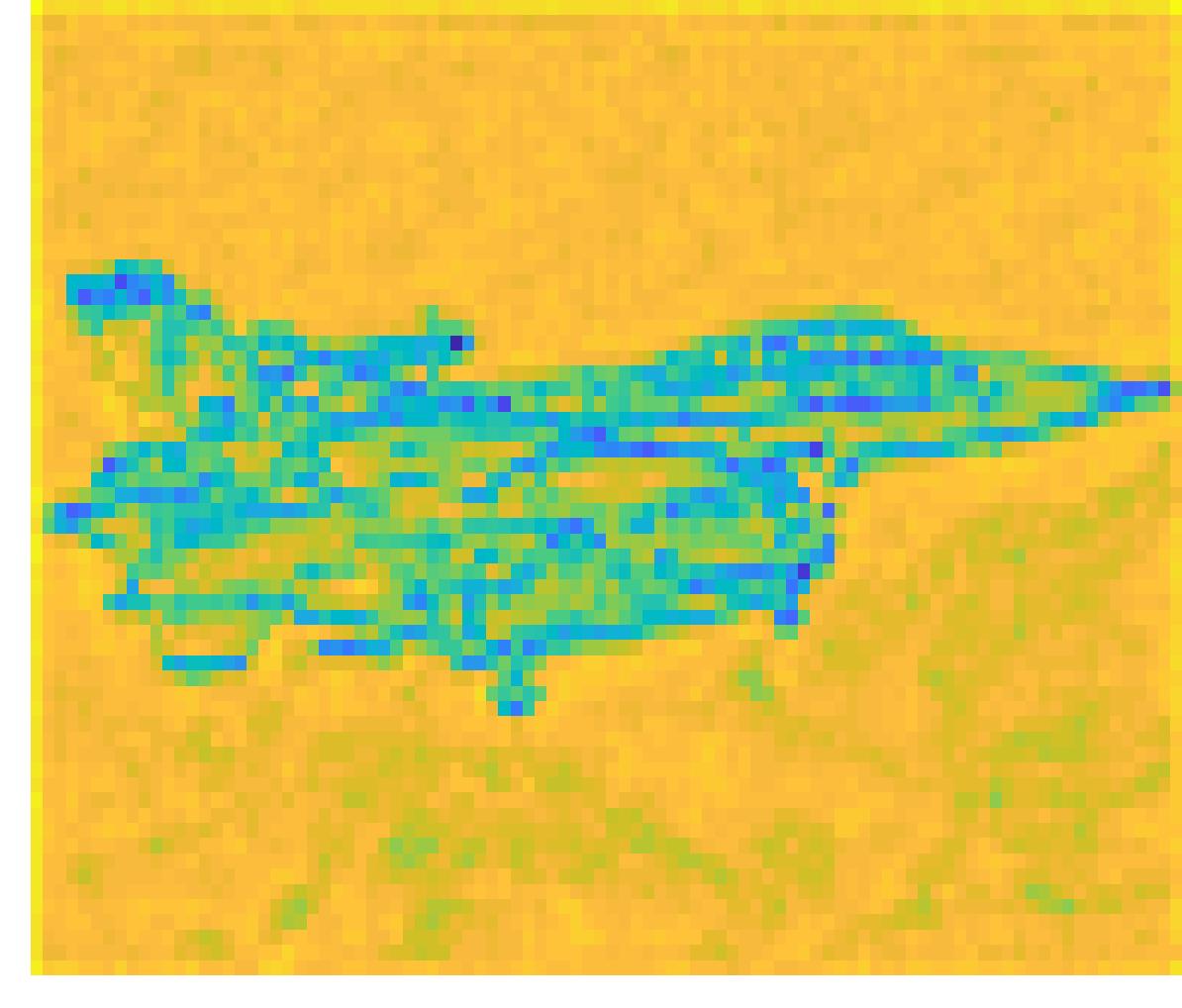}&
     \includegraphics[width=0.2\linewidth]{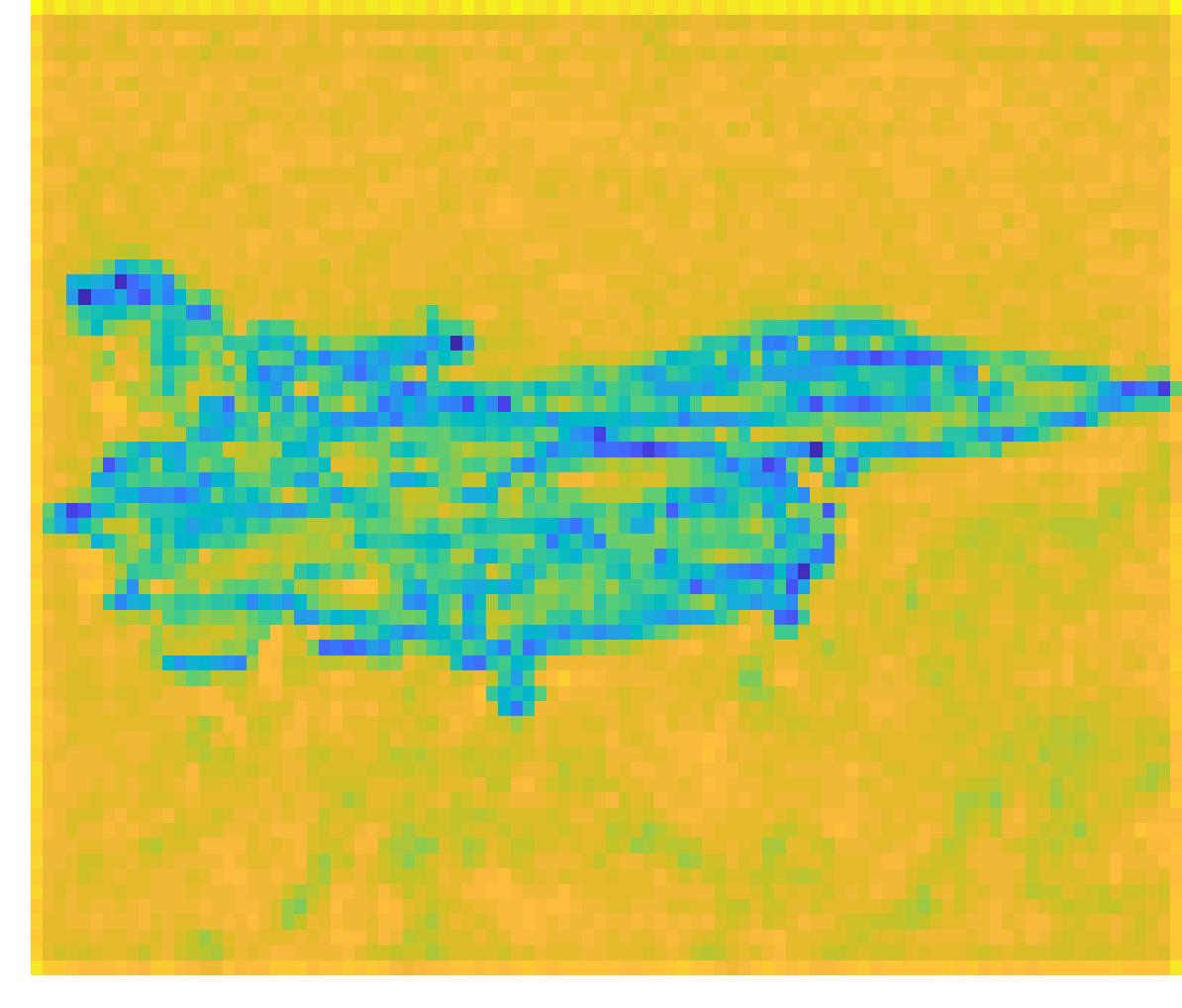}&
     \includegraphics[width=0.2\linewidth]{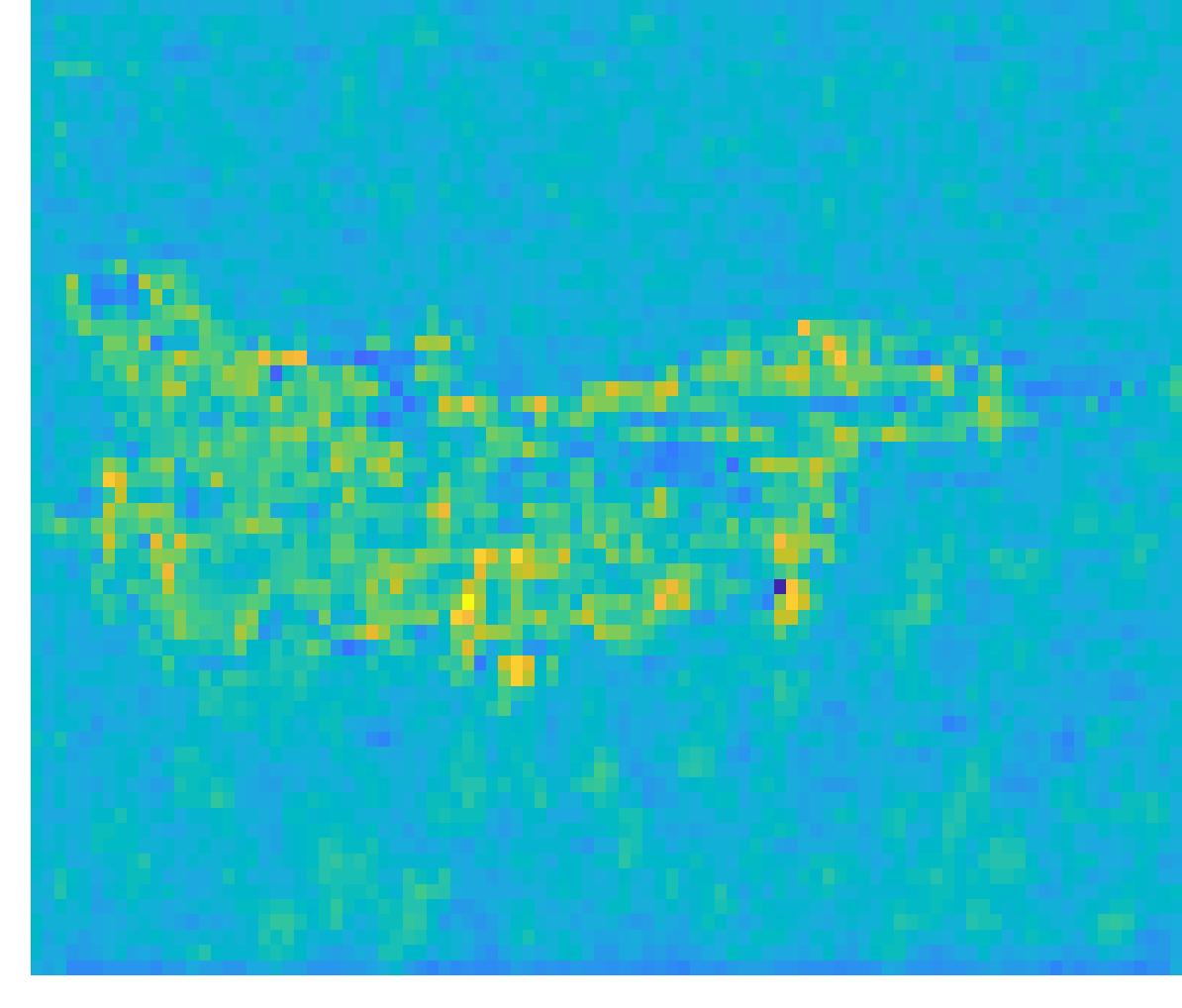}&
     \includegraphics[width=0.2\linewidth]{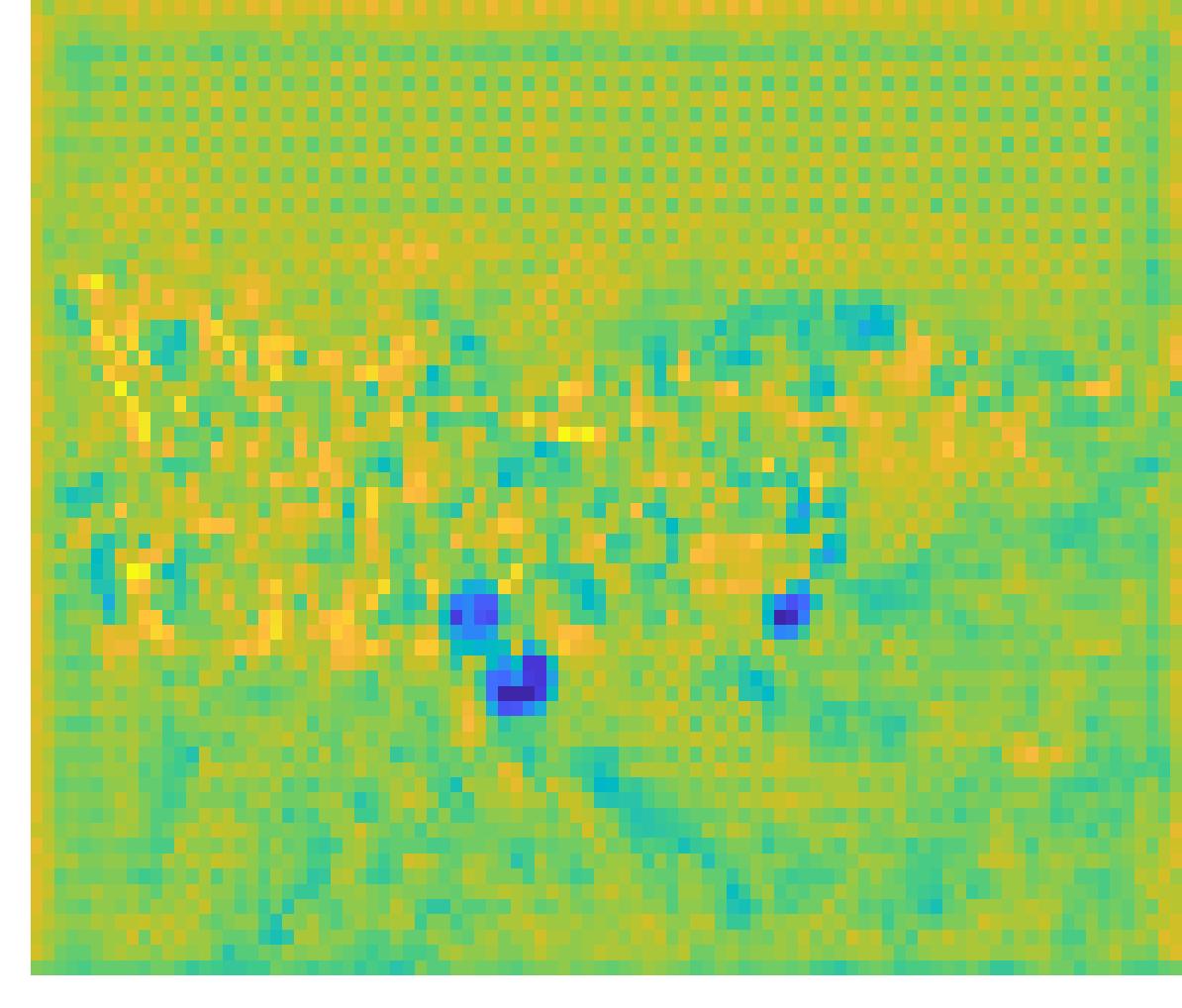}\\[-1ex]     
     \includegraphics[width=0.2\linewidth]{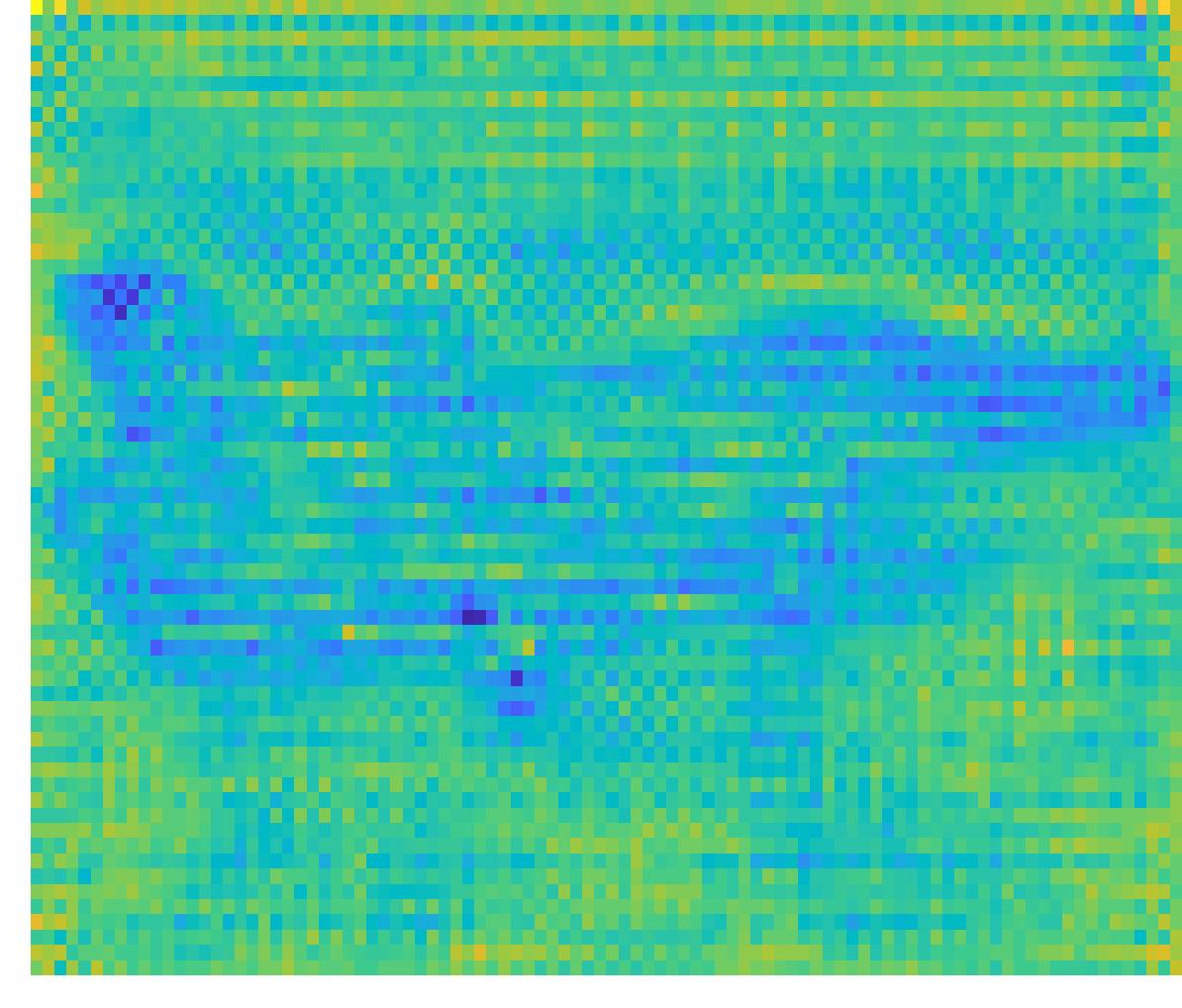}&
     \includegraphics[width=0.2\linewidth]{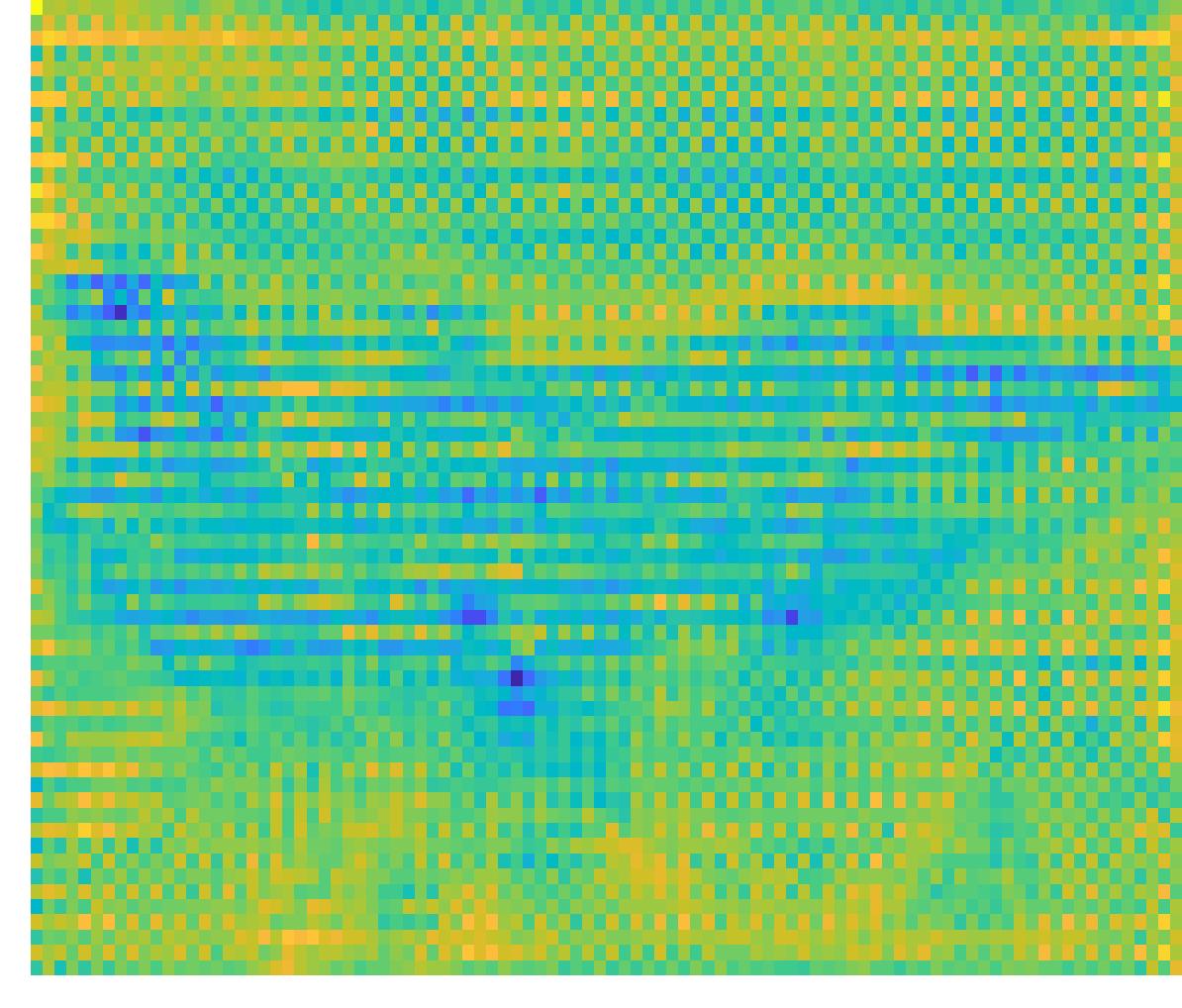}&
     \includegraphics[width=0.2\linewidth]{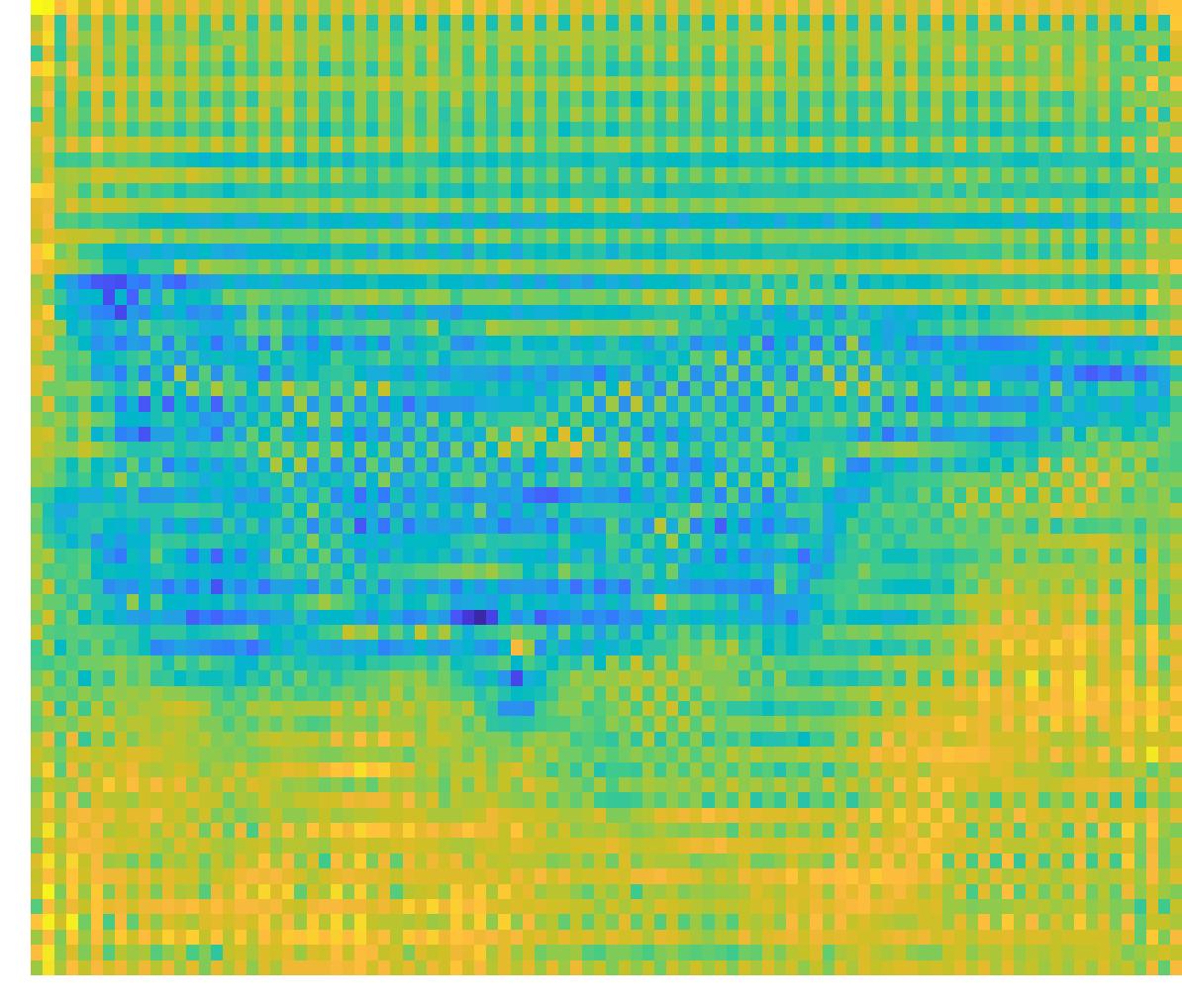}&
     \includegraphics[width=0.2\linewidth]{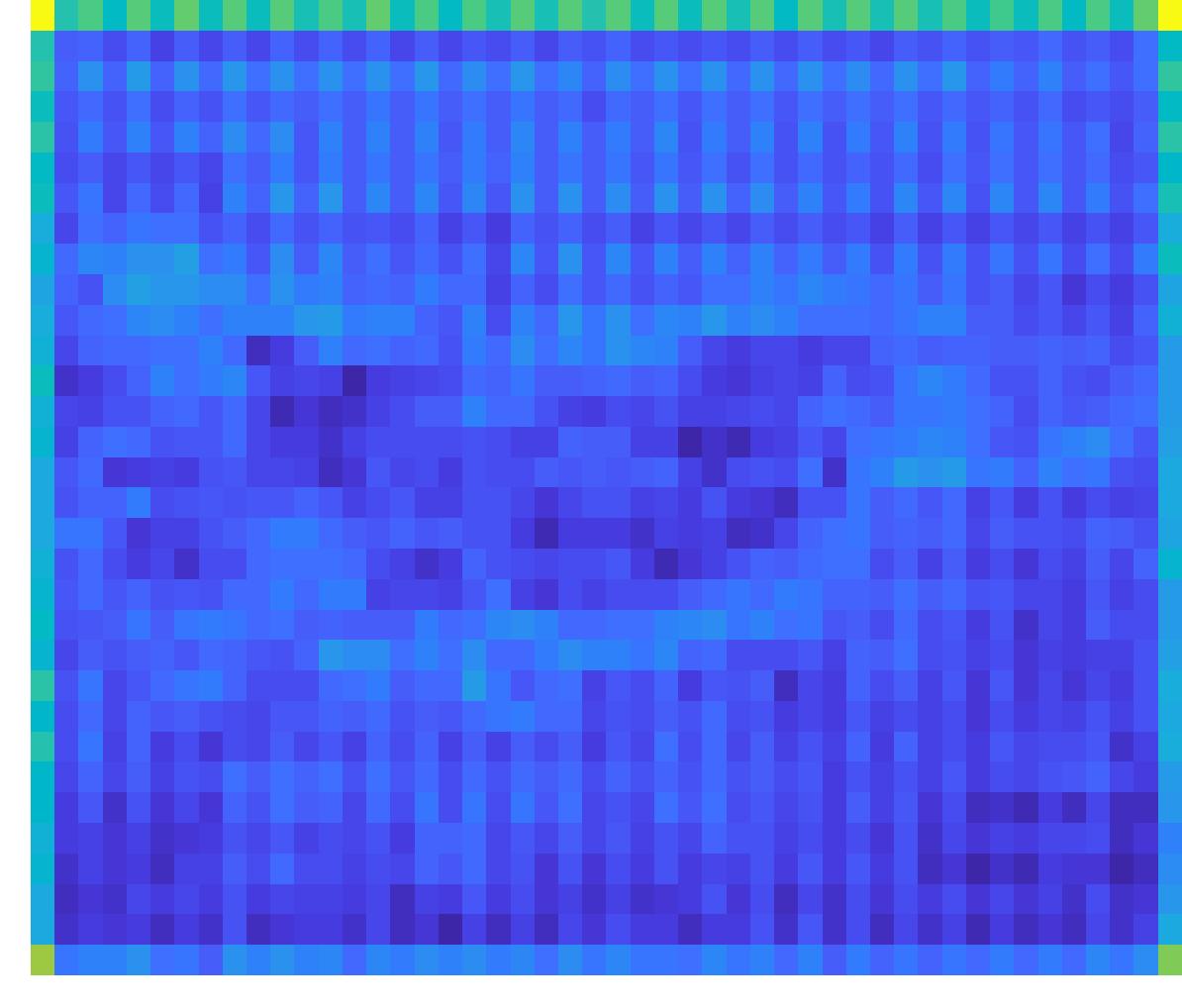}&
     \includegraphics[width=0.2\linewidth]{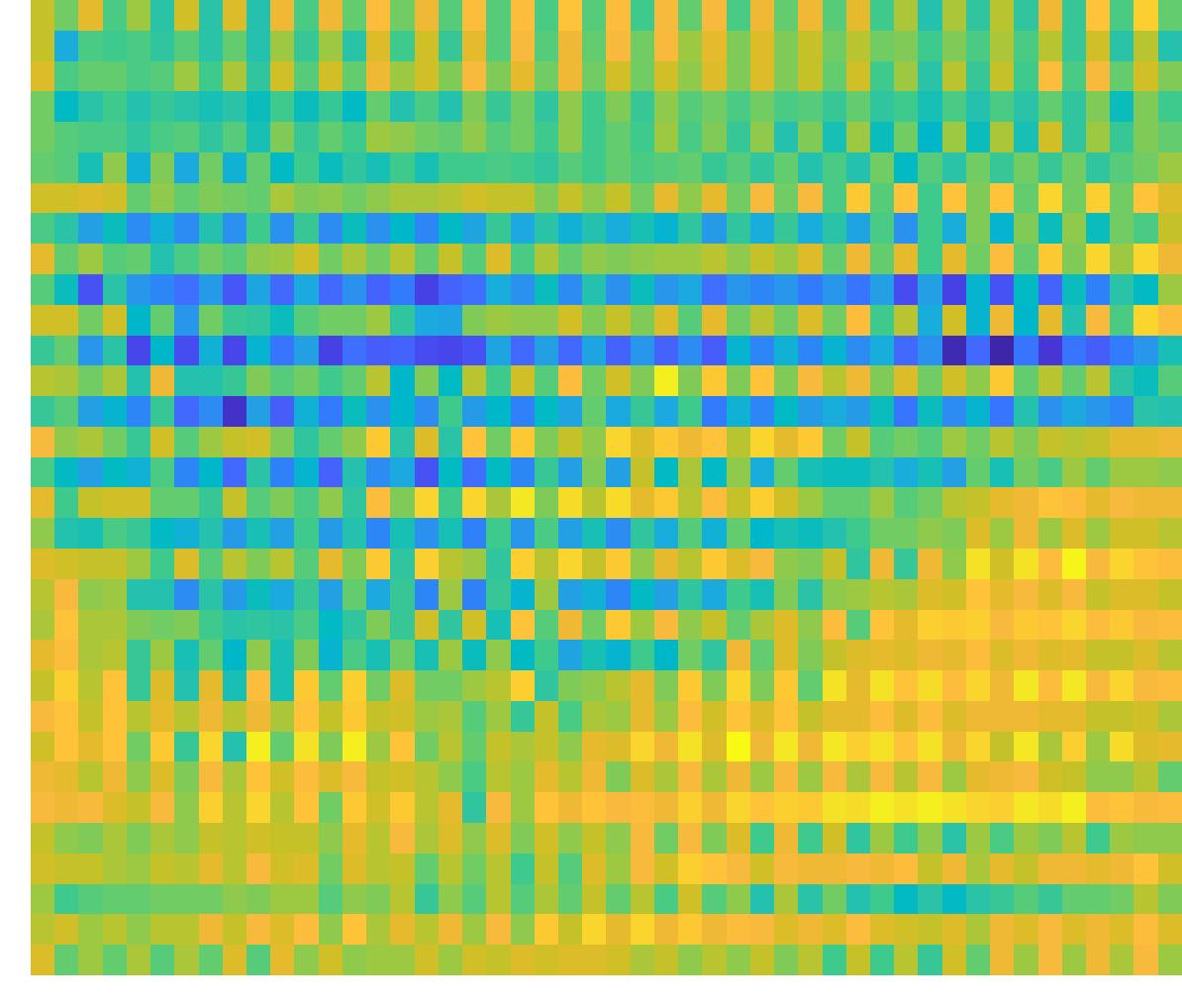}\\[-1ex]     
     \includegraphics[width=0.2\linewidth]{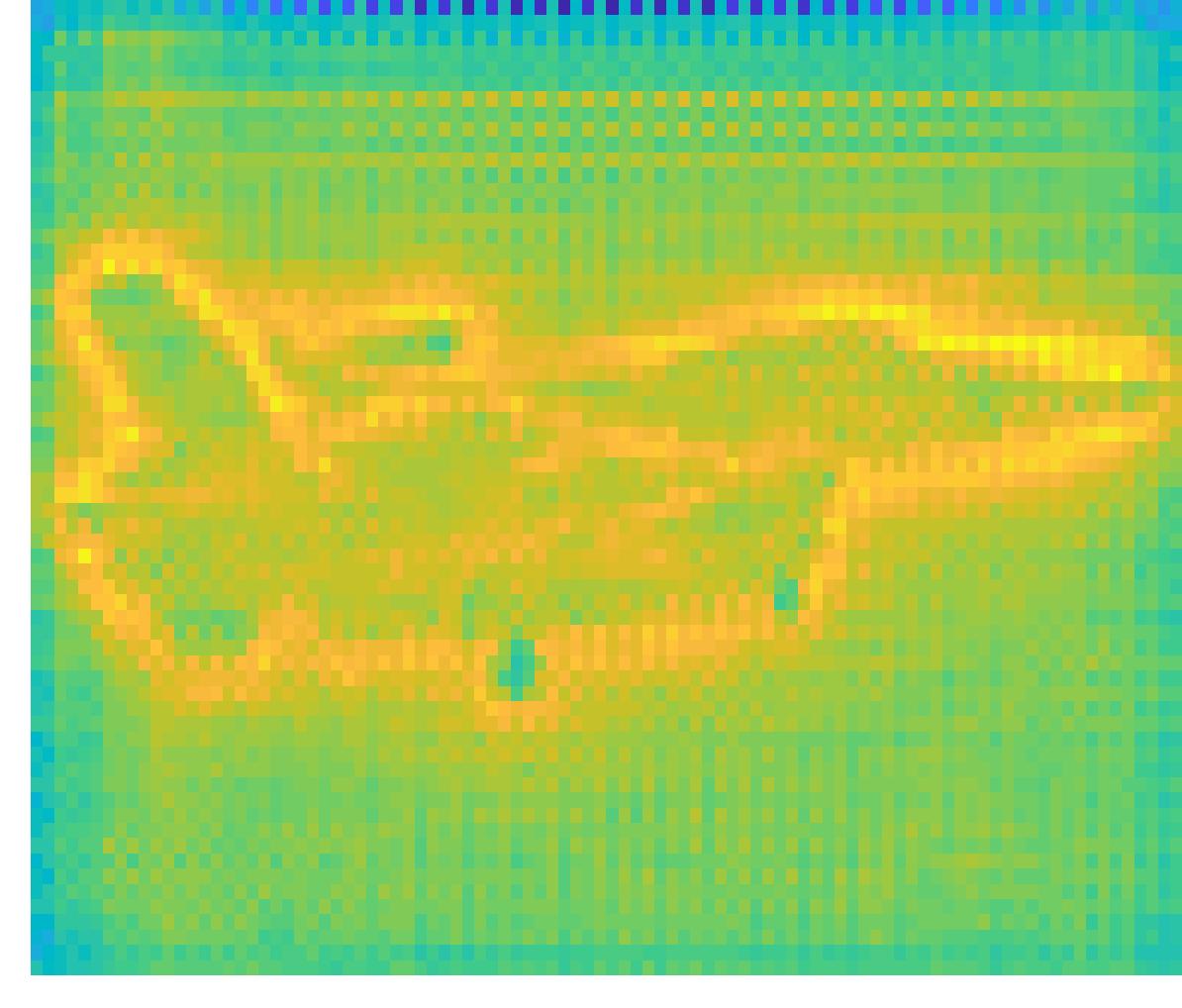}&
     \includegraphics[width=0.2\linewidth]{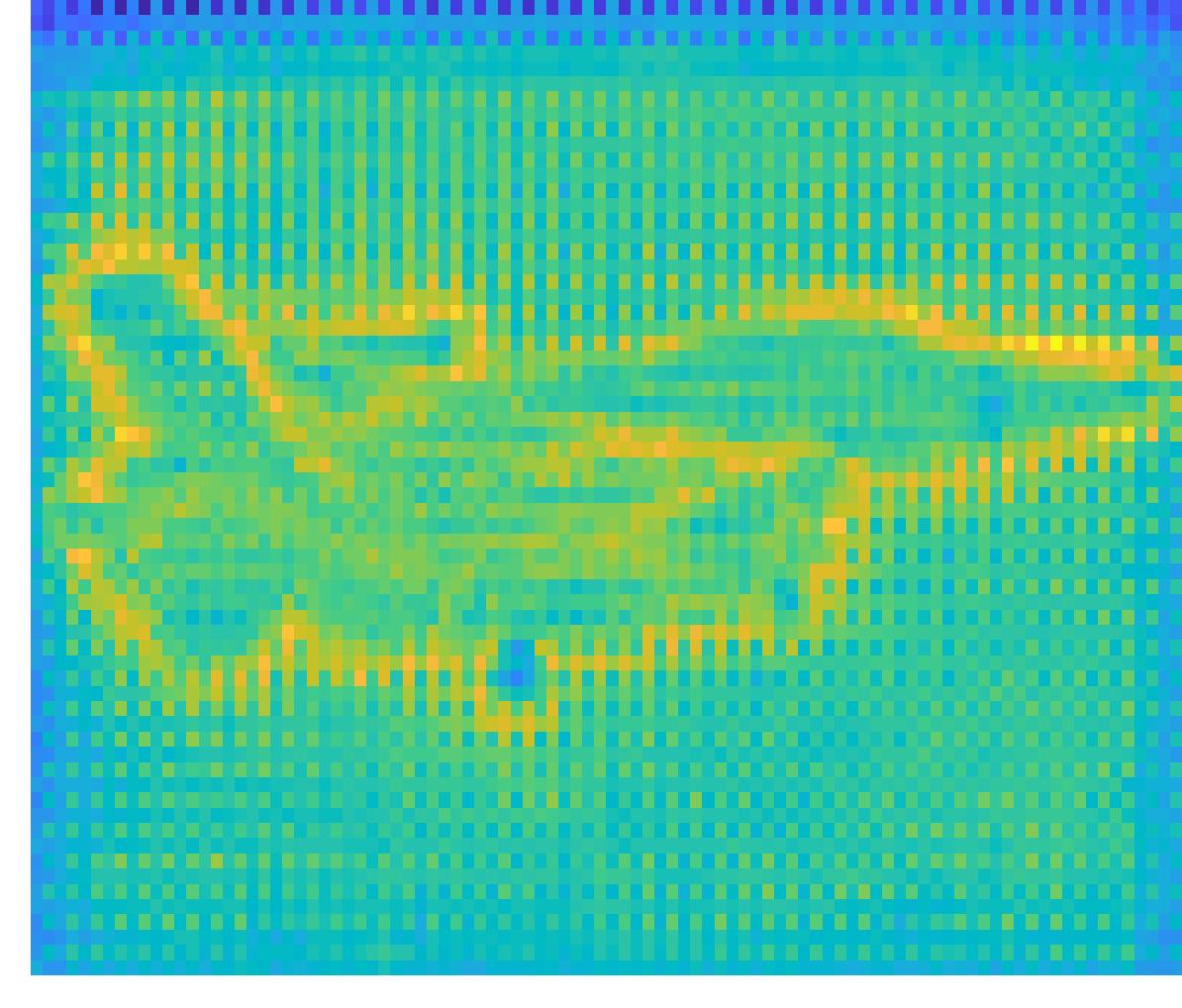}&
     \includegraphics[width=0.2\linewidth]{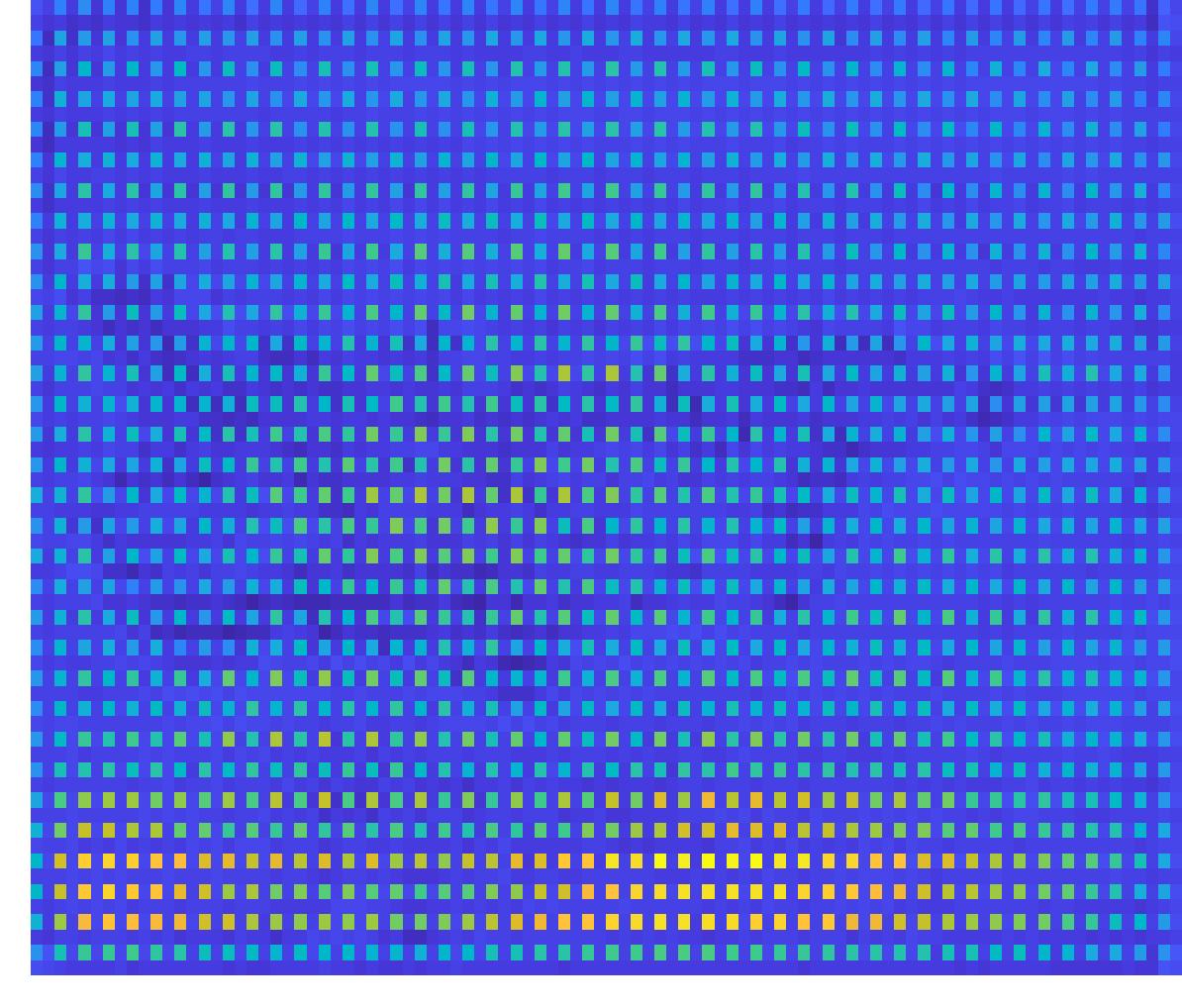}&
     \includegraphics[width=0.2\linewidth]{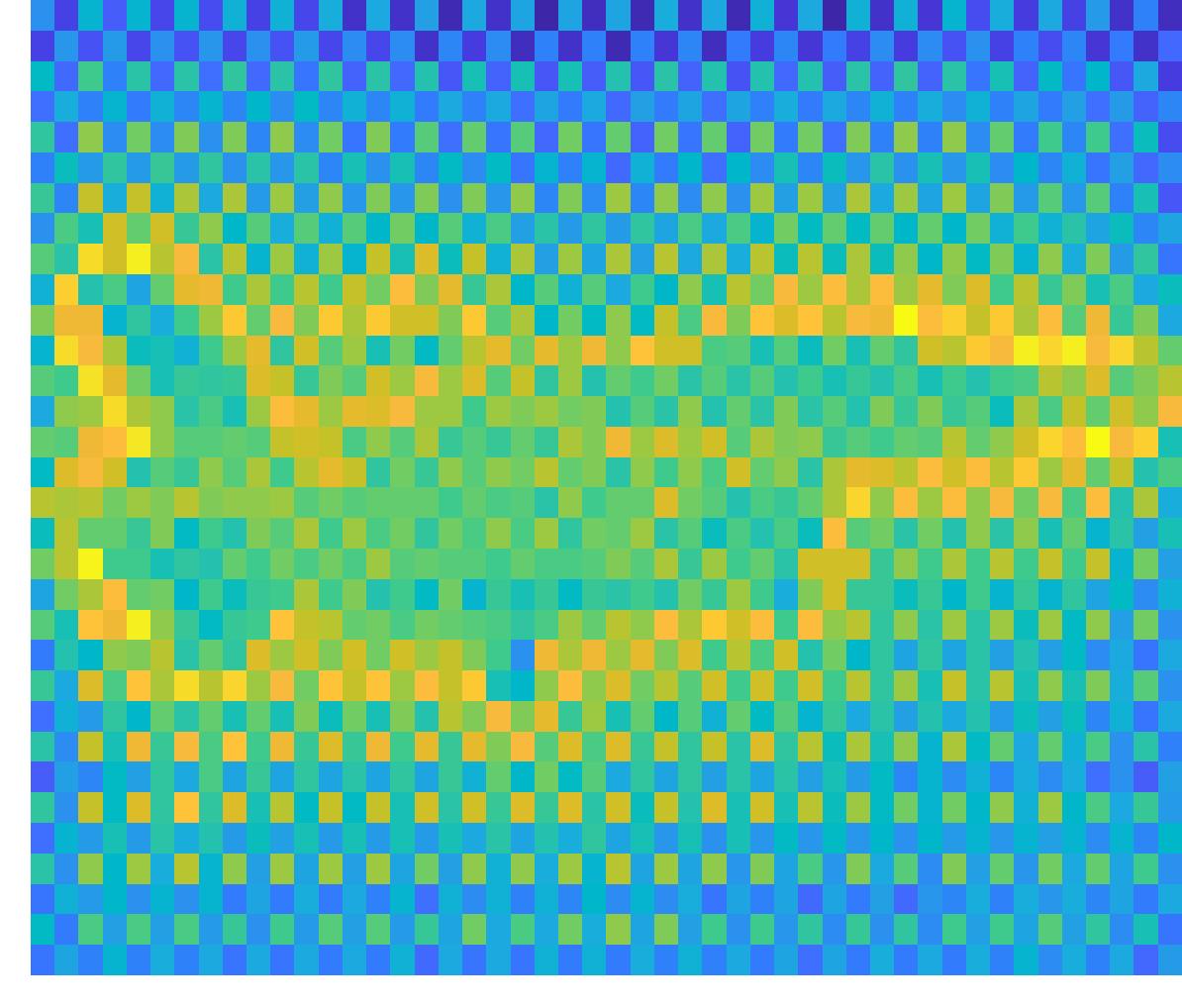}&
     \includegraphics[width=0.2\linewidth]{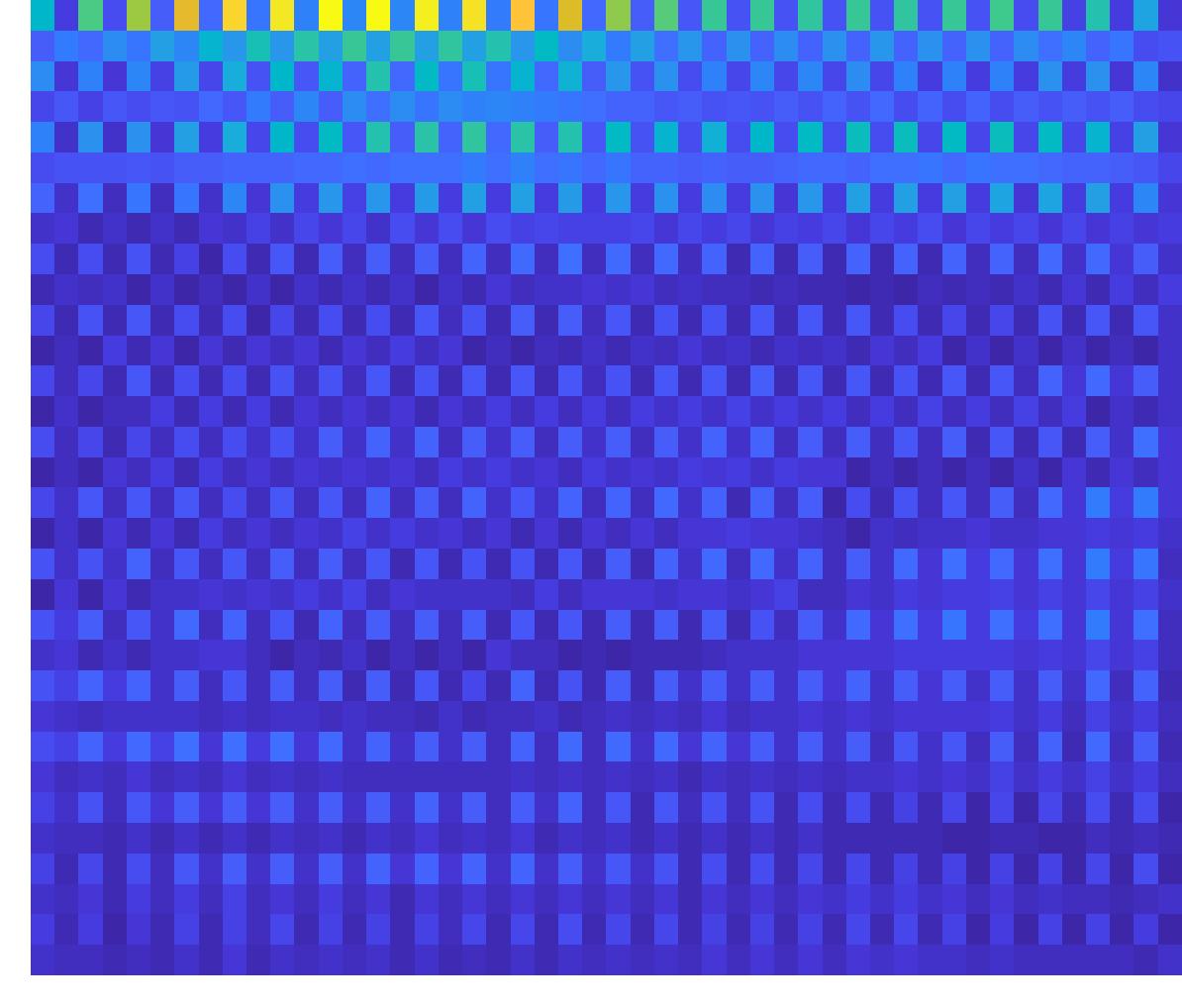}\\[-1ex]     
     \includegraphics[width=0.2\linewidth]{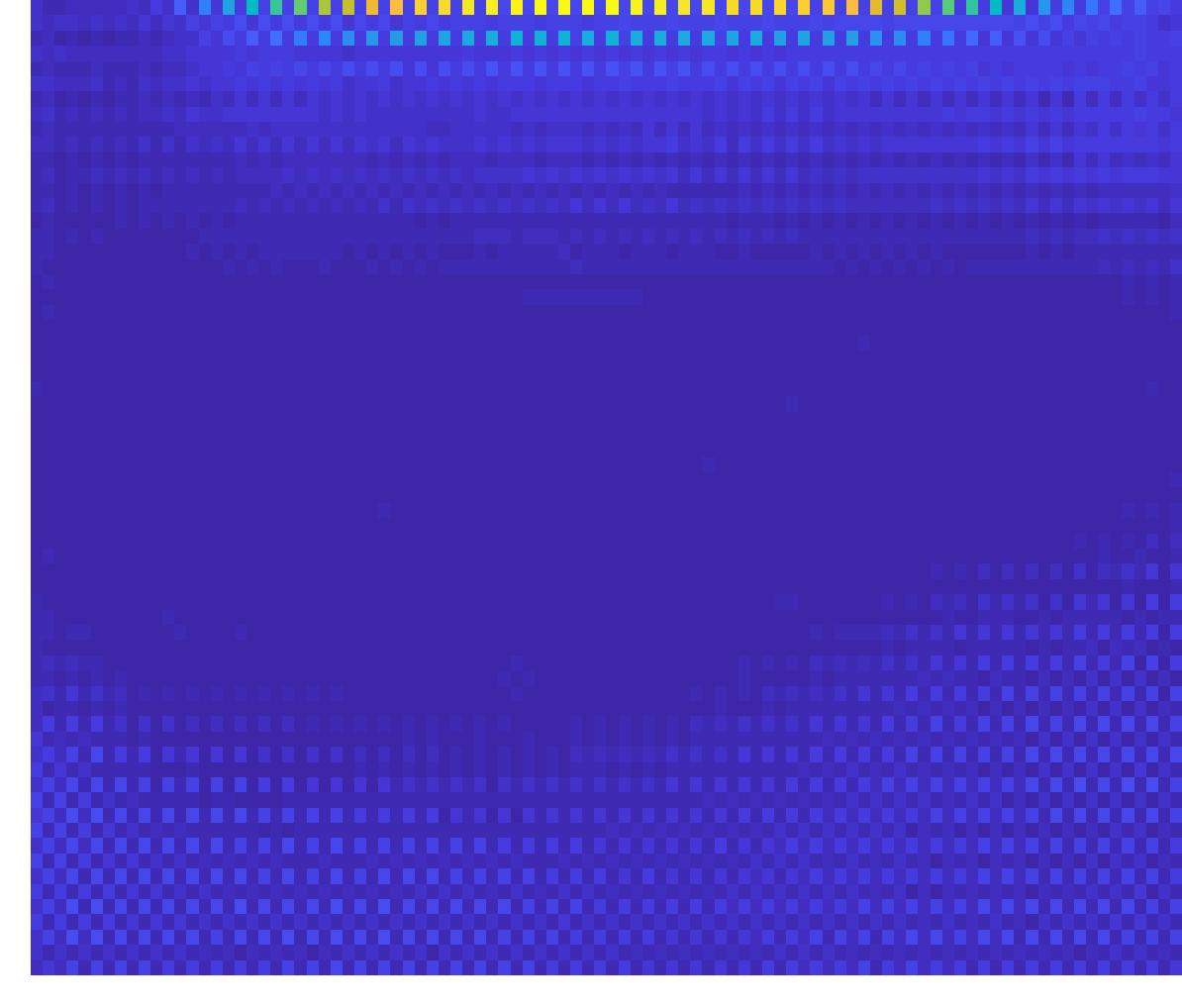}&
     \includegraphics[width=0.2\linewidth]{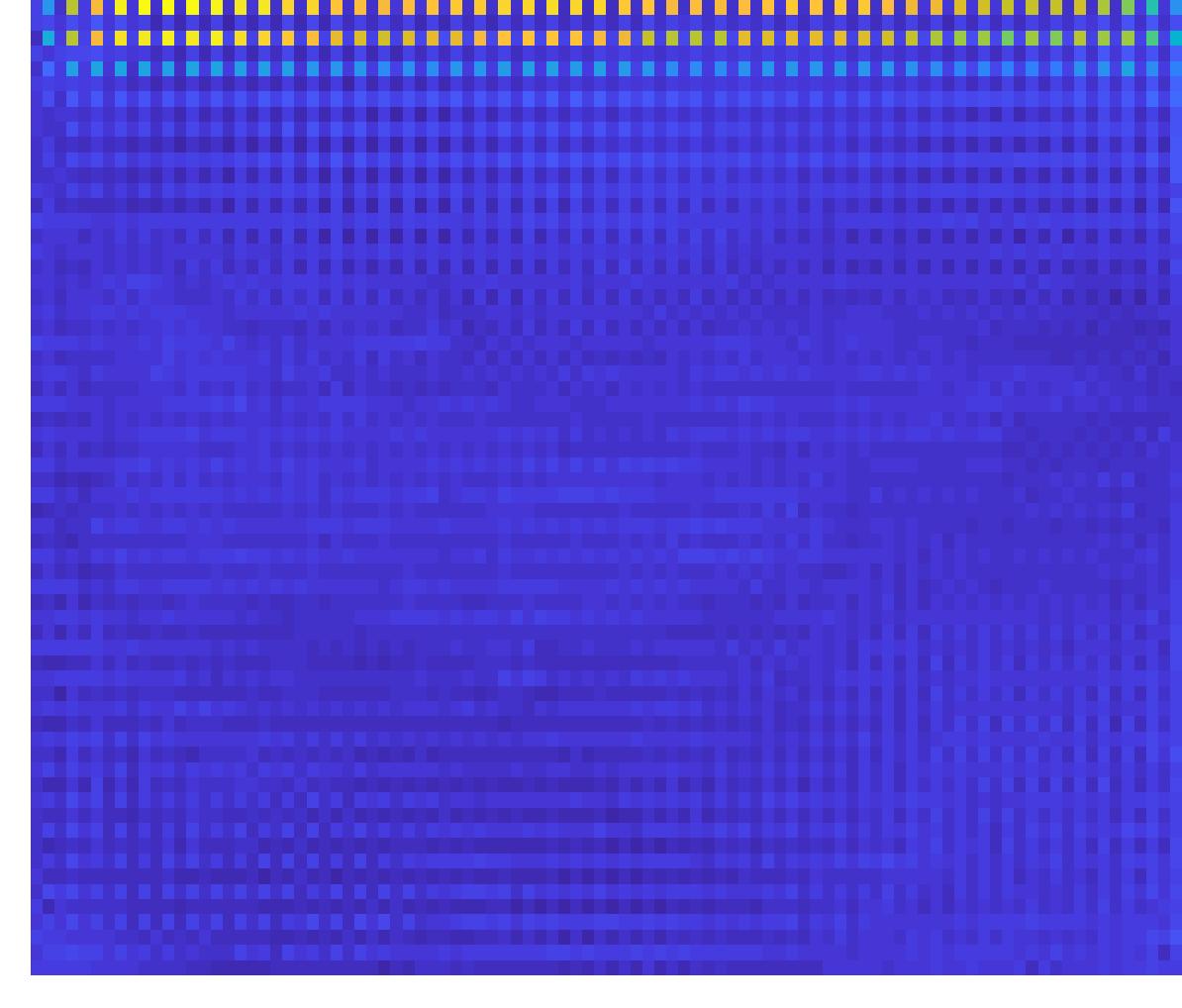}&
     \includegraphics[width=0.2\linewidth]{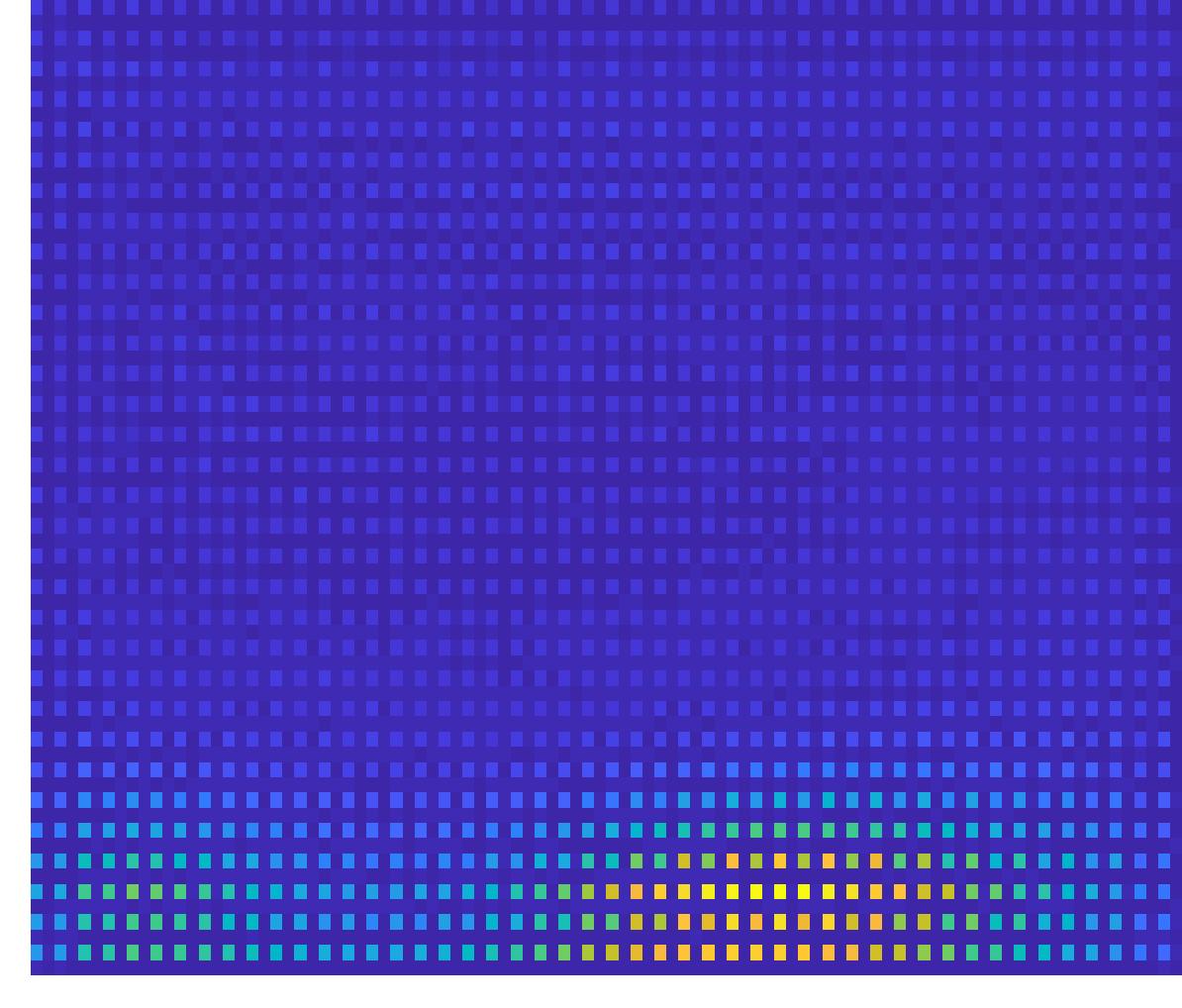}&
     \includegraphics[width=0.2\linewidth]{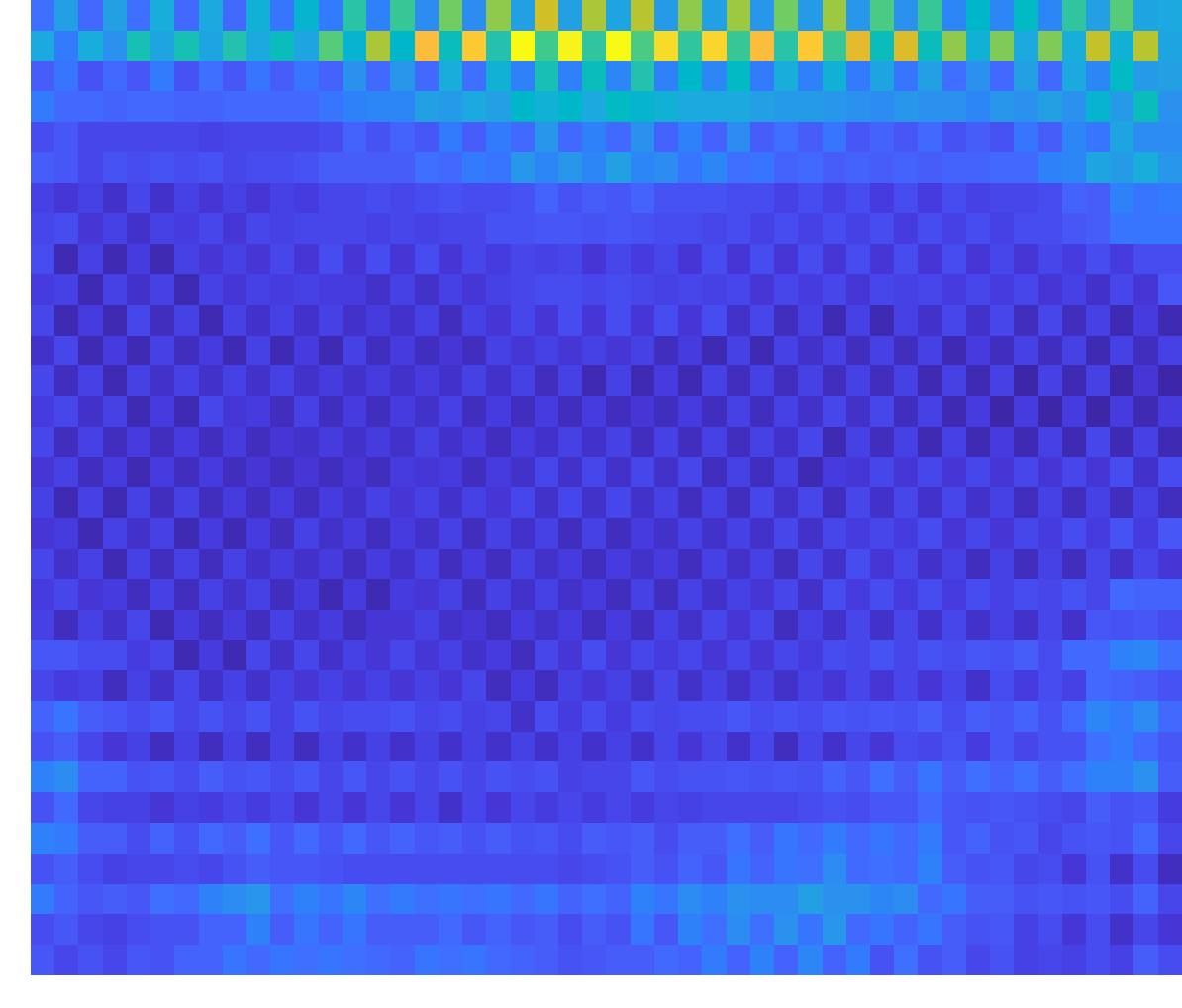}&
     \includegraphics[width=0.2\linewidth]{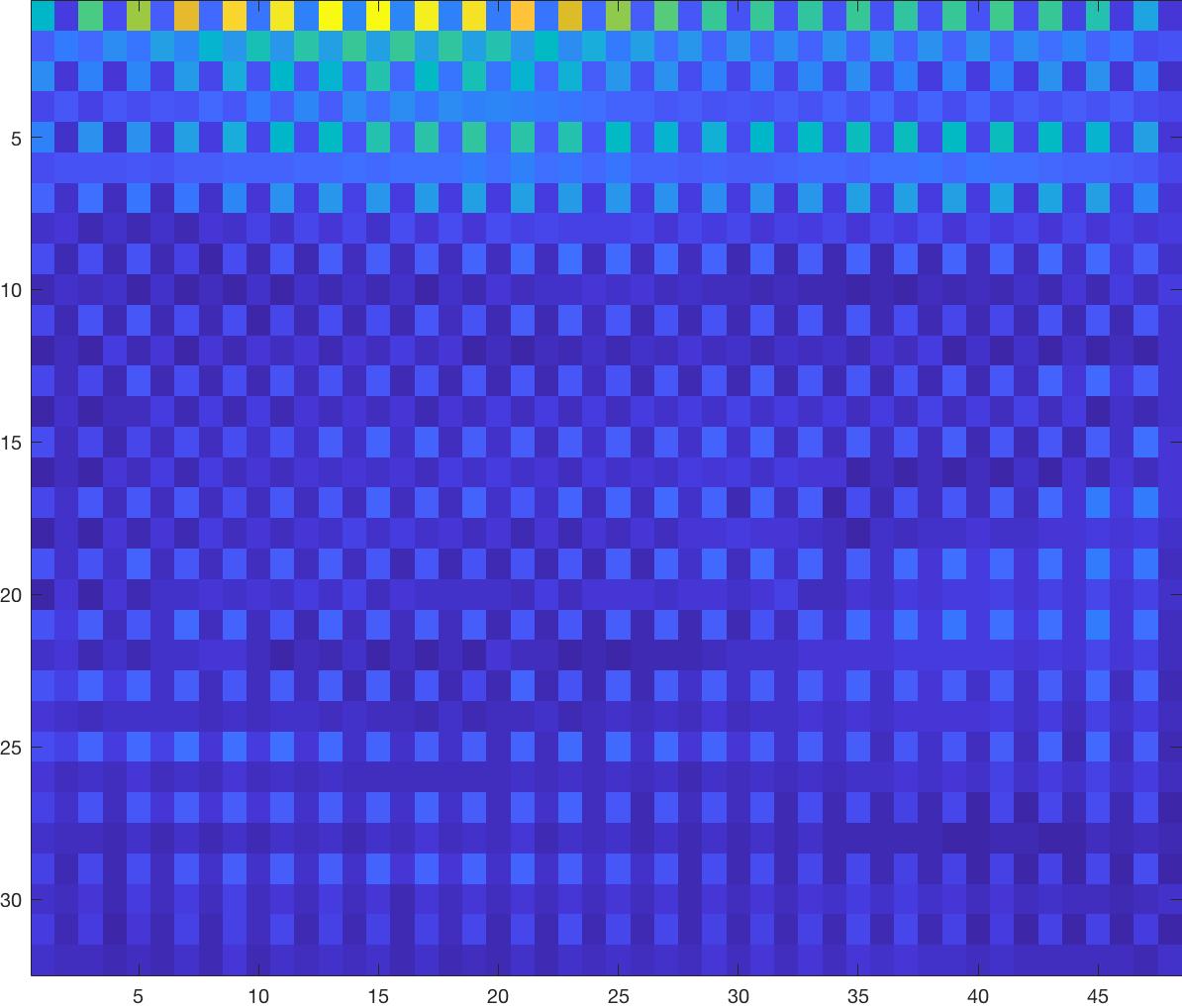}\\[-1ex]     
     \includegraphics[width=0.2\linewidth]{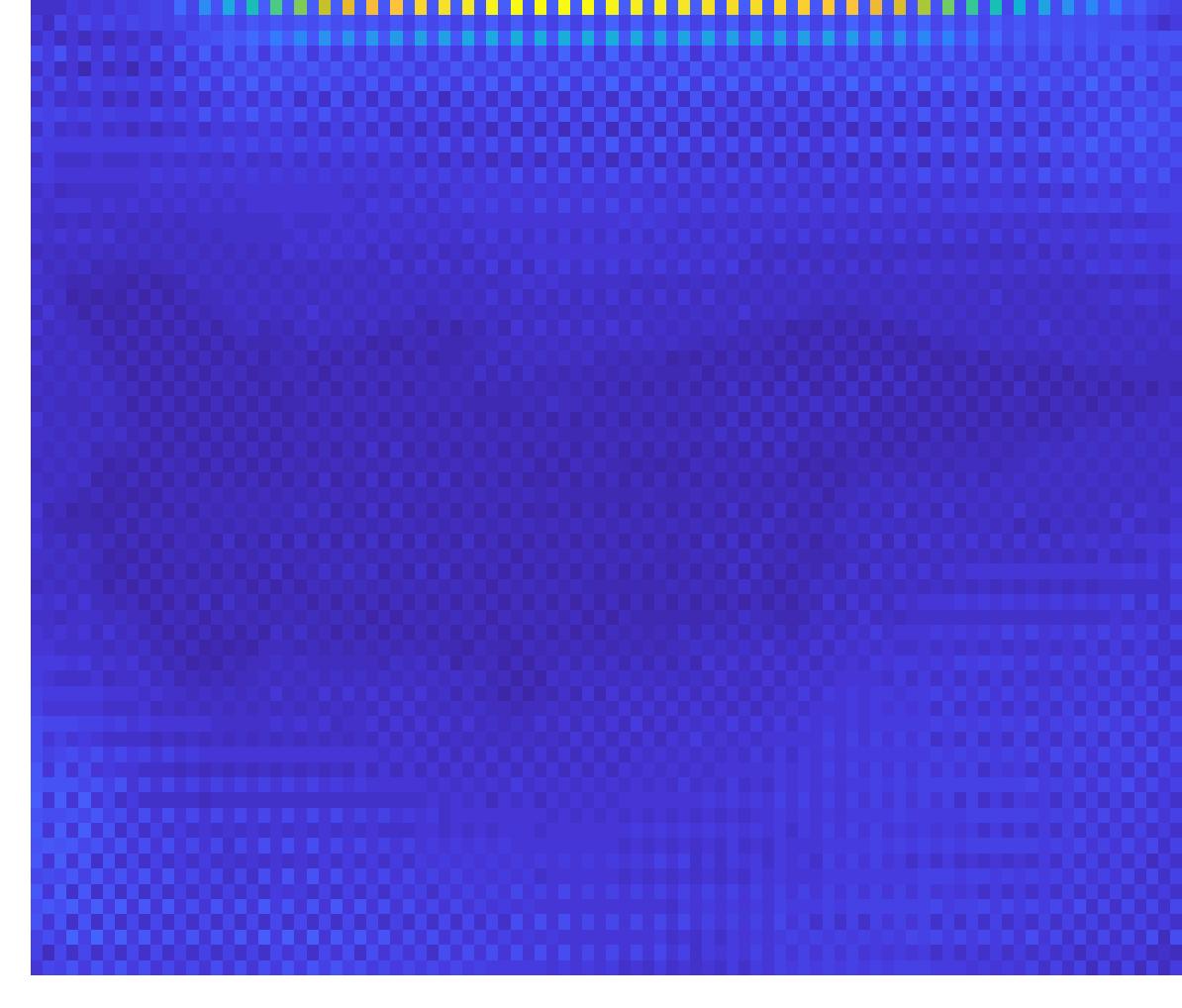}&
     \includegraphics[width=0.2\linewidth]{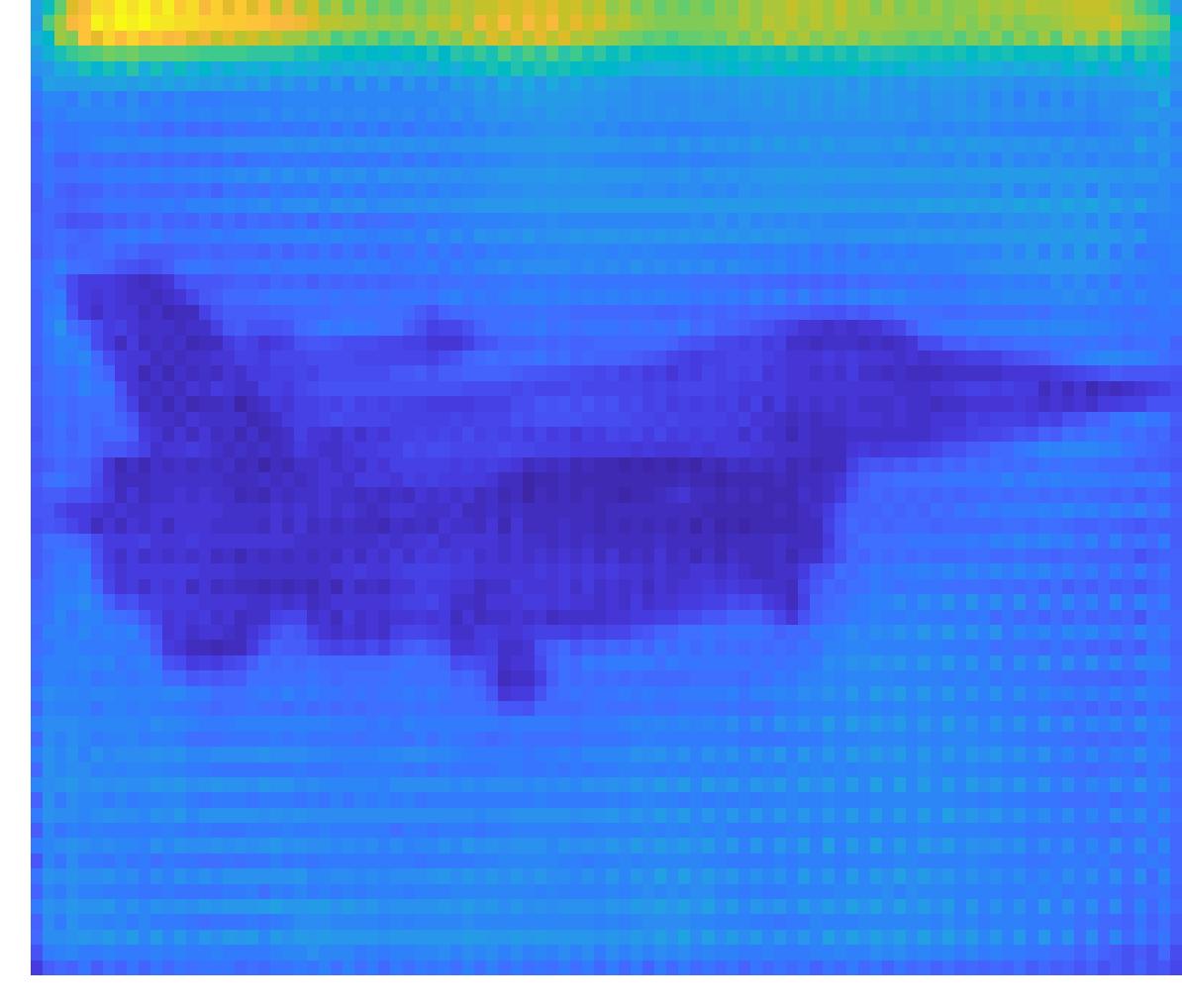}&
     \includegraphics[width=0.2\linewidth]{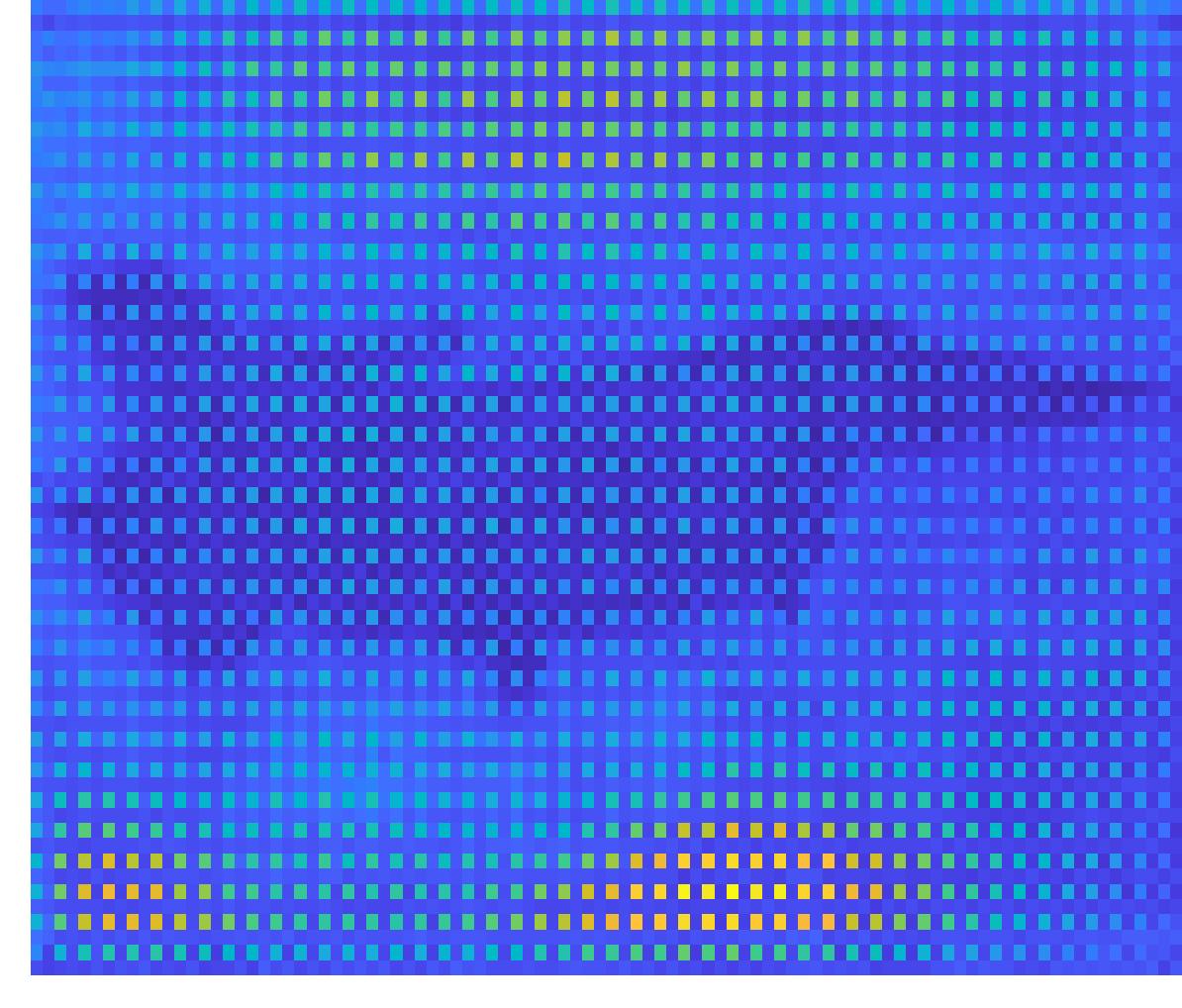}&
     \includegraphics[width=0.2\linewidth]{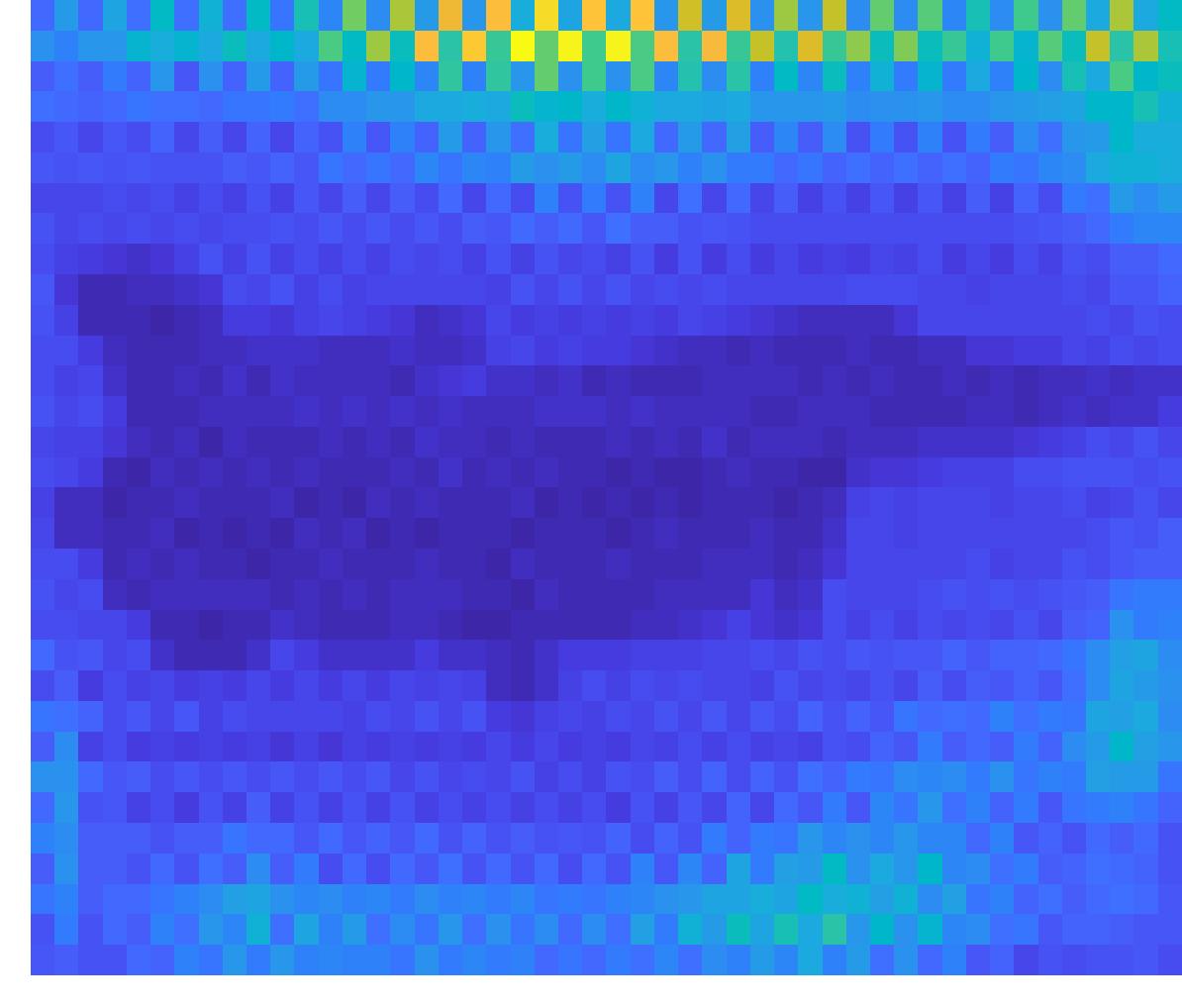}&
     \includegraphics[width=0.2\linewidth]{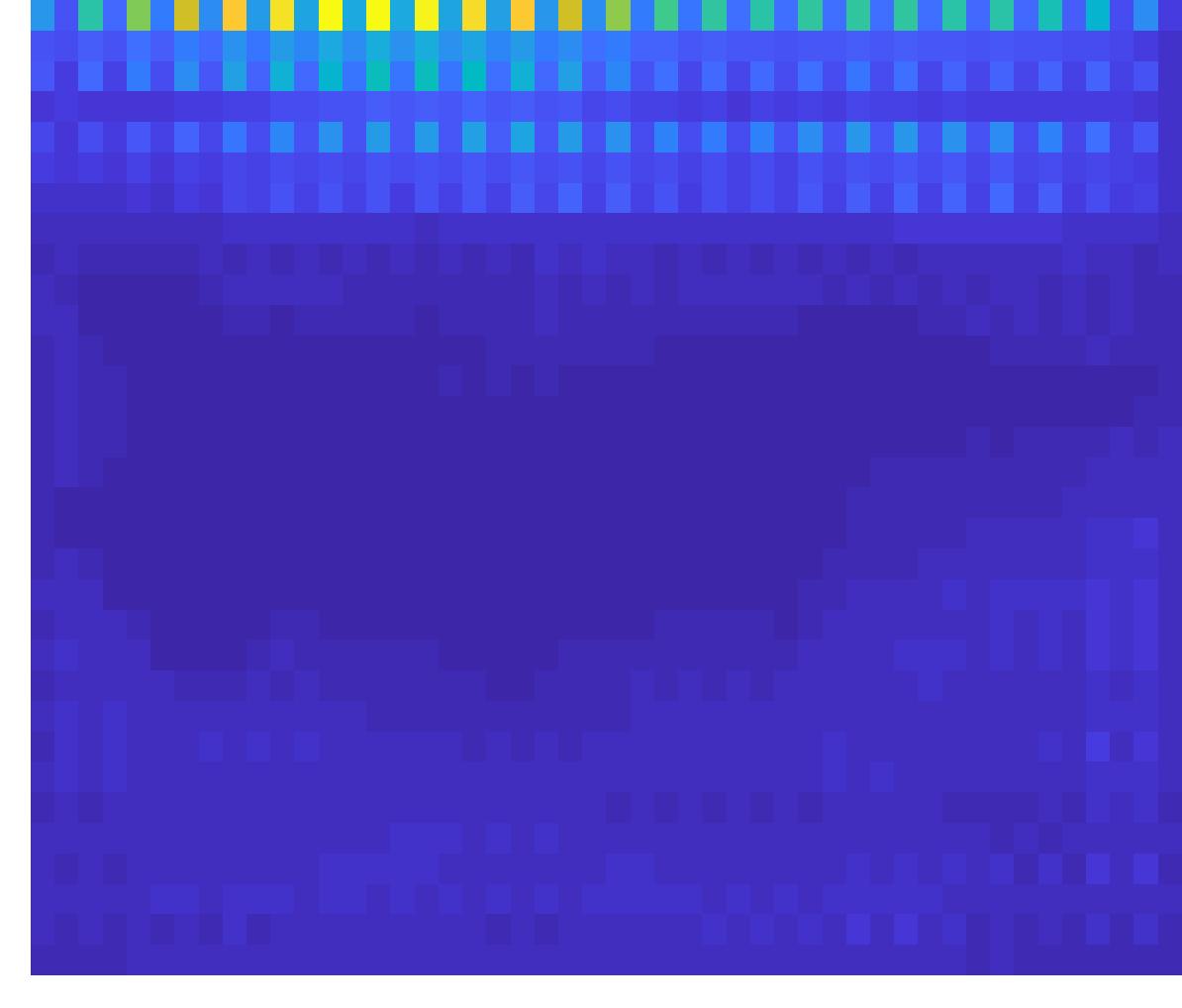}\\[-1ex]     
     \includegraphics[width=0.2\linewidth]{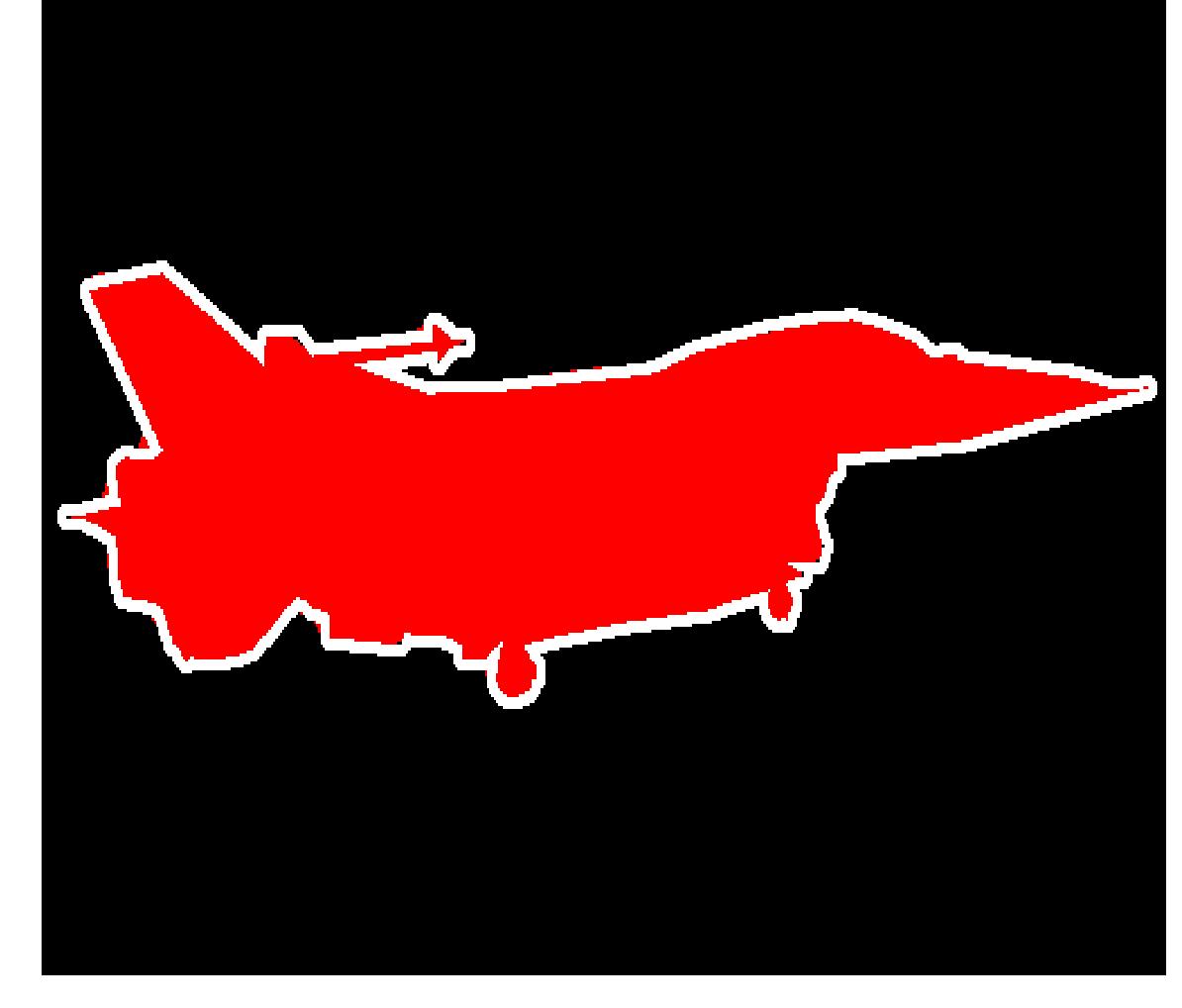}&
     \includegraphics[width=0.2\linewidth]{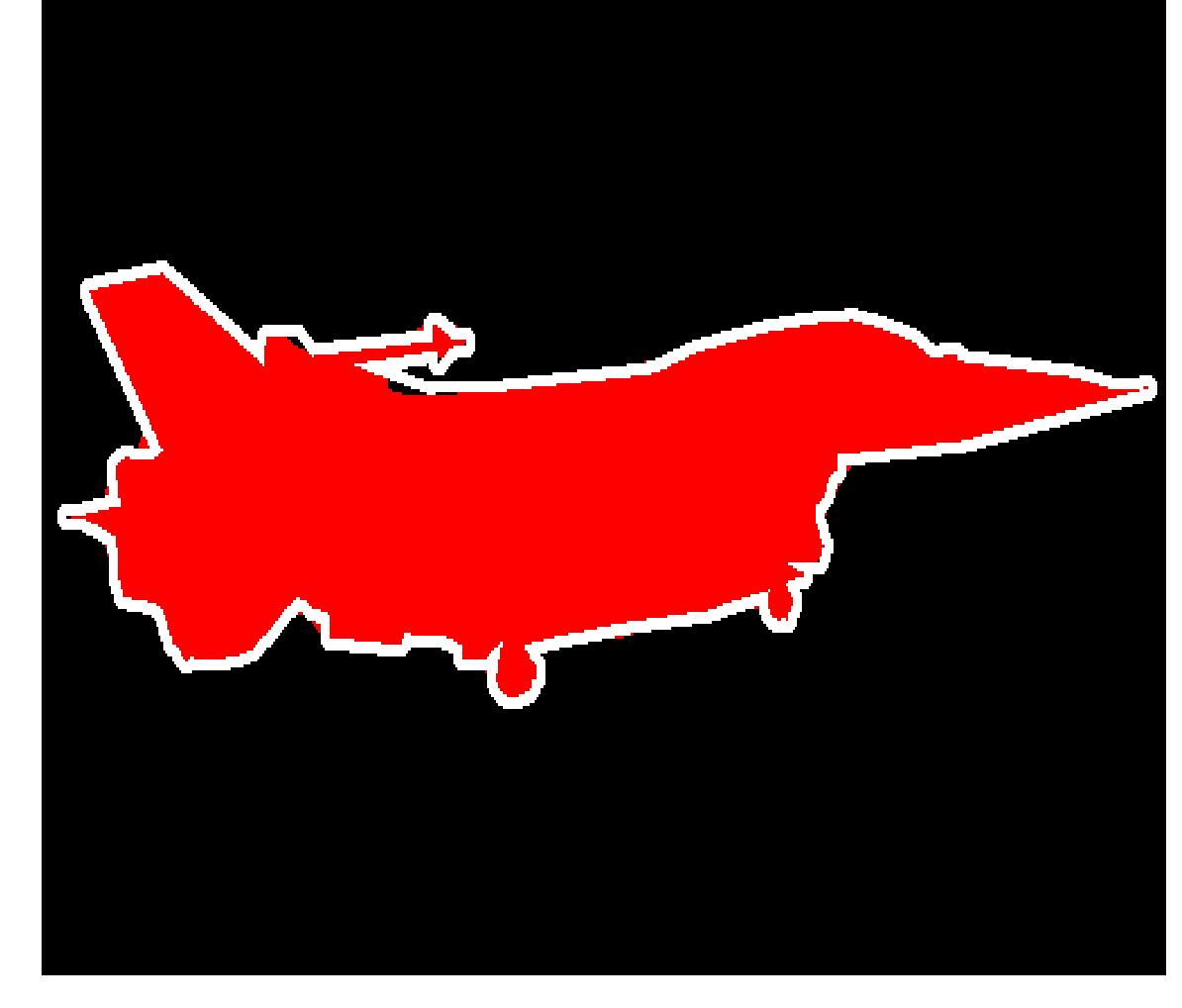}&
     \includegraphics[width=0.2\linewidth]{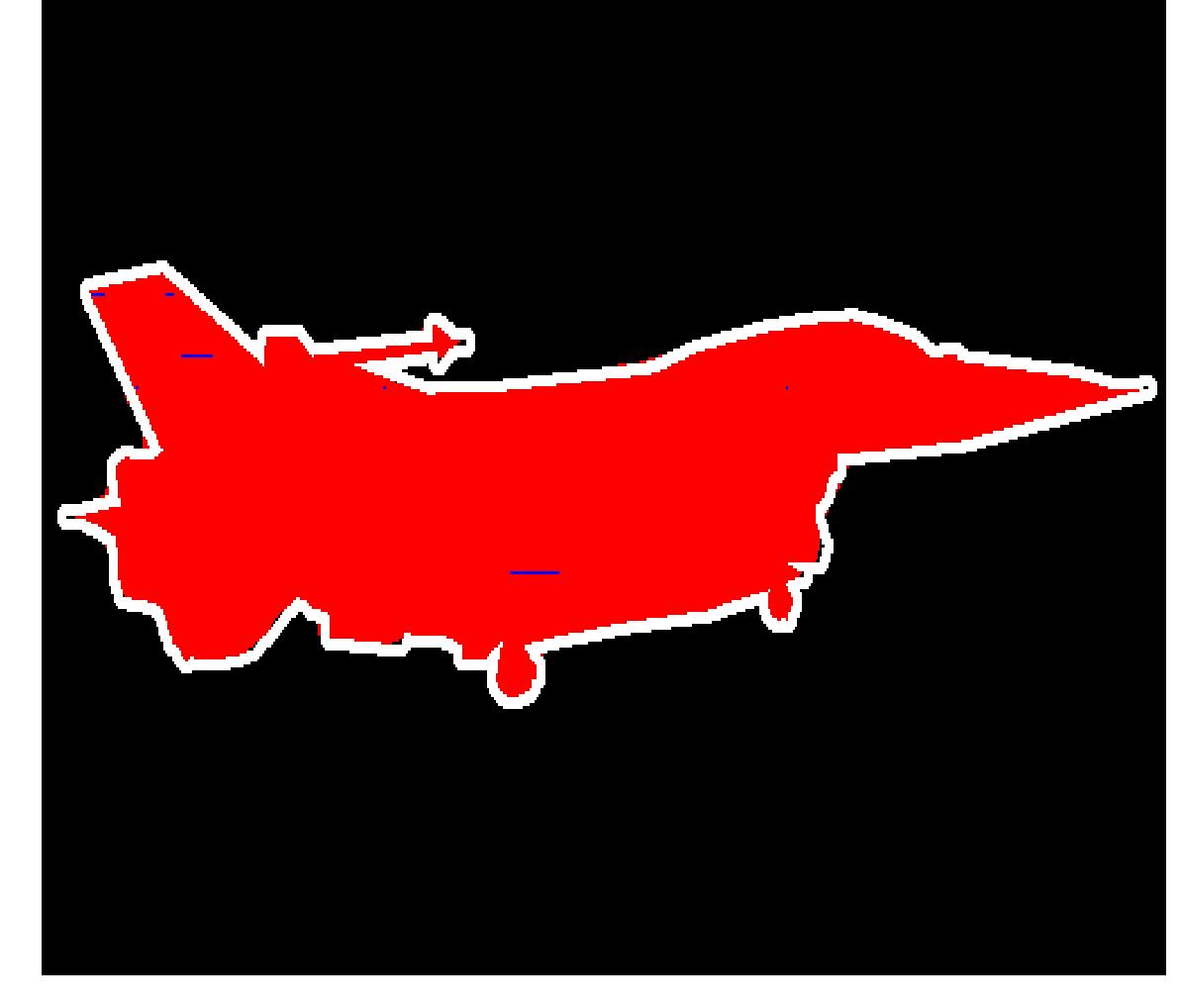}&
     \includegraphics[width=0.2\linewidth]{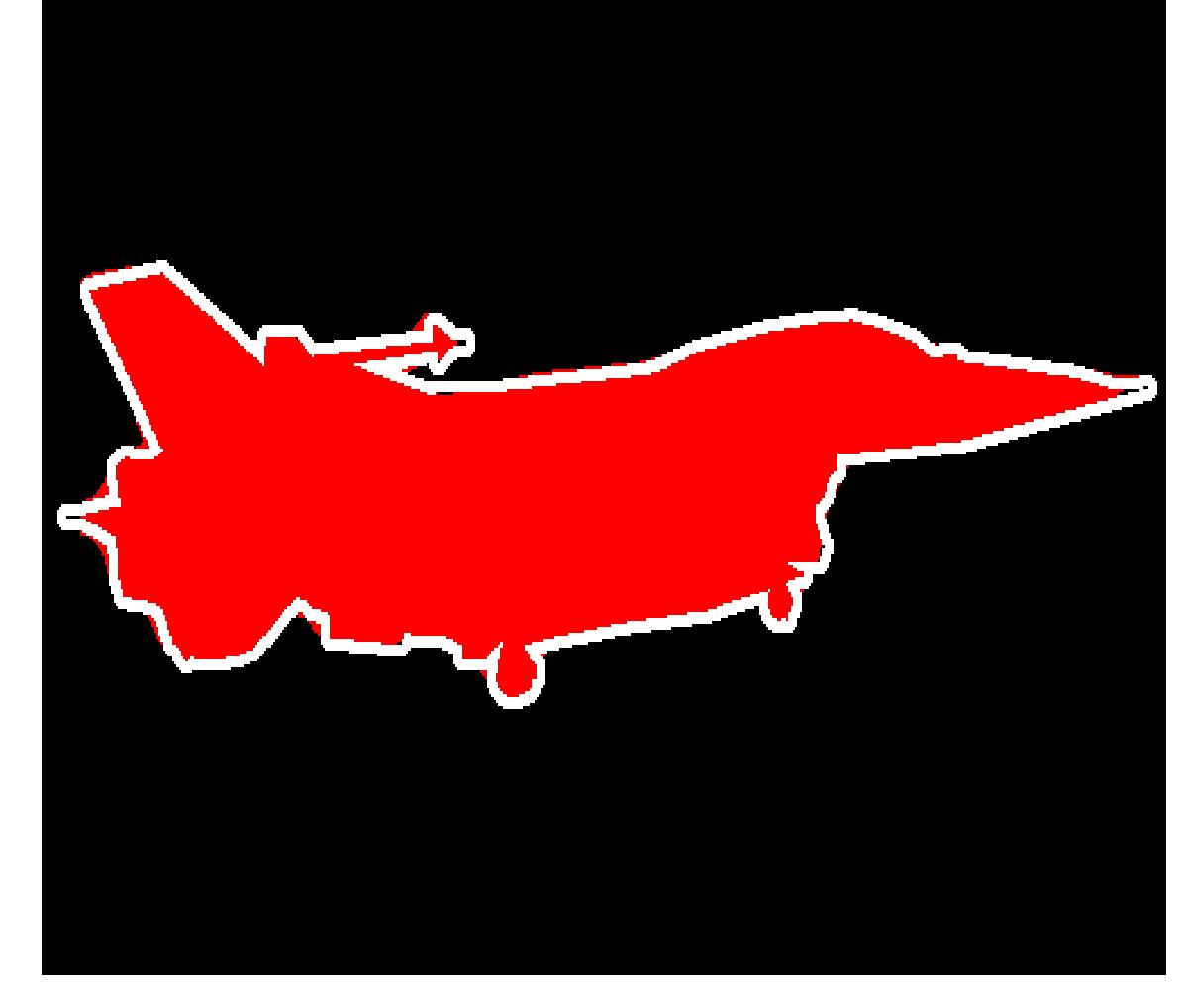}&
     \includegraphics[width=0.2\linewidth]{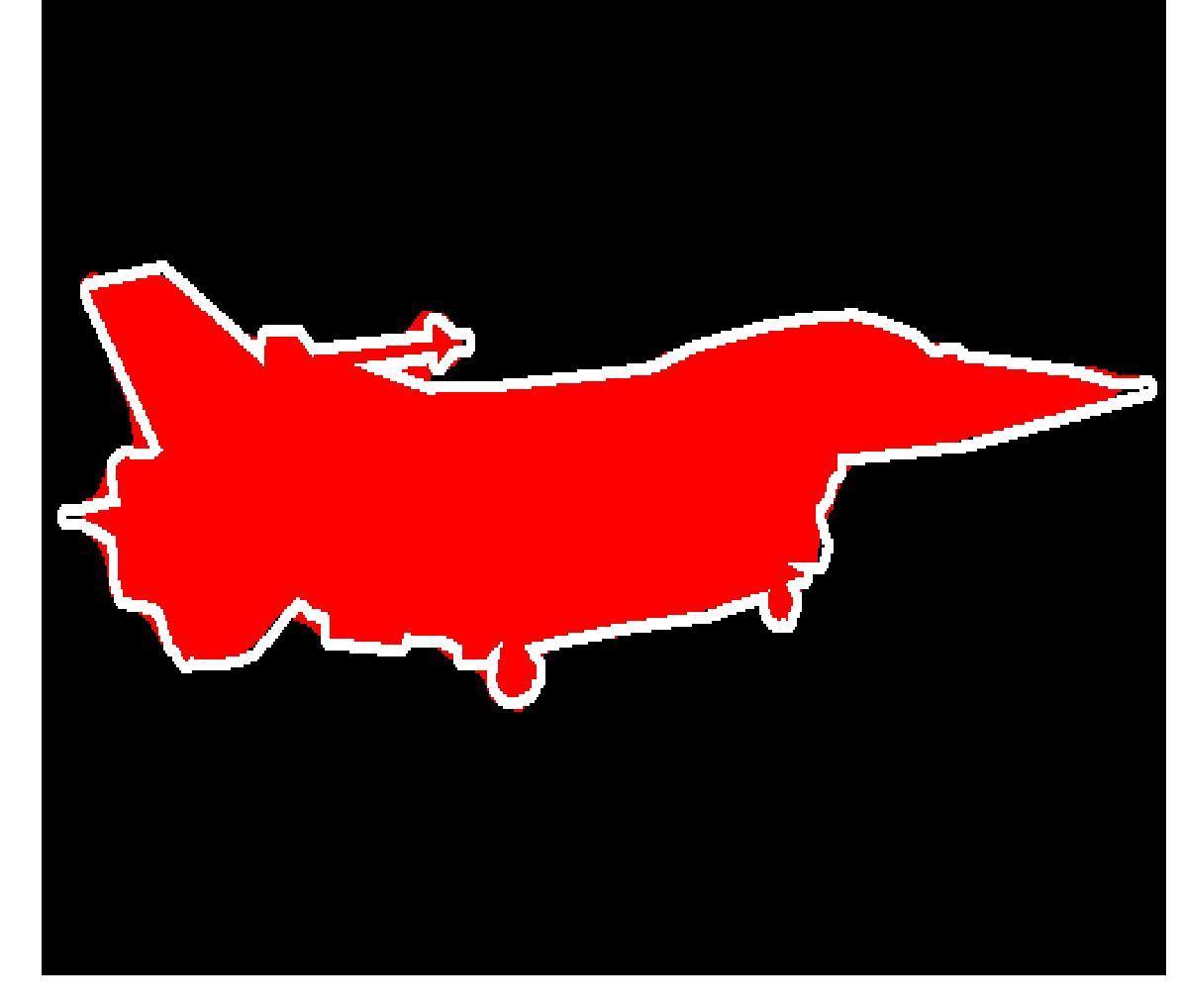}\\
   \end{tabular}
\caption{Activation map recorded at the end of each level of the dilated SUNet for an example input image of an `Aeroplane'. The activation map with total highest magnitude were selected from among all feature map outputs at the corresponding layer. \textbf{Top} to \textbf{Bottom}: output at end of level $1-6$ followed by classification output. The level 6 output is simply a prediction map before bilinear interpolation.}
\label{fig:act}
\end{figure}

Figure~\ref{fig:act} shows the activation map recorded at the end of each level (as indicated in figure~\ref{fig:seg}) for an example input image of an ``Aeroplane.'' As noted earlier, the inclusion of strided convolutions instead of multigrid dilations leads to noisy feature maps (see col 3; rows 4-6). The addition of de-gridding layers serves to produce a coherent prediction map at the output (see col 2; row 6).

\section{Discussion and Conclusion}
The fundamental structure of conventional bottom-up classification networks limits their efficacy on secondary tasks involving pixel-level localization or classification. To overcome this drawback, a new network architecture, stacked u-nets (SUNets), is discussed in this paper. SUNets leverage the information globalization power of u-nets in a deeper network architecture that is capable of handling the complexity of natural images.
 SUNets perform exceptionally well on semantic segmentation tasks while achieving fair performance on ImageNet classification. 

There are several directions for future research that may improve upon the performance achievable using a simple SUNet.  
It may be advantageous to replace each convolution block by their corresponding depthwise separable convolution~\cite{chollet2017xception}, as done in~\cite{dai2017coco,deeplabv3xception,zhang2017shufflenet,howard2017mobilenets}. 
The inclusion of post-hoc context~\cite{zhao2017pyramid,chen2017rethinking} or decoder networks~\cite{deeplabv3xception} on top of SUNets may also help.
Given the huge margin of improvement over ResNet models for semantic segmentation tasks, it is obvious to extend SUNets to object detection tasks.
Finally, as suggested in the paper, rigorous hyper-parameter search and numerical analysis~\cite{li2017visualizing} on SUNets may yield better generic as well as task-specific models.

\bibliographystyle{splncs}
\bibliography{paper}
\end{document}